\pdfoutput=1
\documentclass[11pt]{article}

\usepackage[final]{ACL2023}

\usepackage{times}
\usepackage{latexsym}
\usepackage{graphicx}
\usepackage{subfigure}
\usepackage{enumitem}

\usepackage[T1]{fontenc}
\usepackage[utf8]{inputenc}
\usepackage{amsmath}

\usepackage{microtype}

\usepackage{inconsolata}
\usepackage{booktabs}
\usepackage{multirow}
\usepackage{bm}
\usepackage{supertabular, longtable}

\title{Pre-Training to Learn in Context}
\author{
  Yuxian Gu$^{1,2,*}$, 
  Li Dong$^2$,
  Furu Wei$^2$,
  Minlie Huang$^{1,\dagger}$\\
\small $^1$The CoAI Group, DCST, Institute for Artificial Intelligence, State Key Lab of Intelligent Technology and Systems, \\
\small Beijing National Research Center for Information Science and Technology,
Tsinghua University, Beijing 100084, China \\
\small $^2$ Microsoft Research \\
  \small \texttt{guyx21@mails.tsinghua.edu.cn}, \ \ \ \texttt{\{lidong1,fuwei\}@microsoft.com}\\
  \small \texttt{aihuang@tsinghua.edu.cn}\\
 }

\begin{document}
\maketitle
\begin{abstract}

In-context learning, where pre-trained language models learn to perform tasks from task examples and instructions in their contexts, has attracted much attention in the NLP community. However, the ability of in-context learning is not fully exploited because language models are not explicitly trained to learn in context. To this end, we propose PICL (\textbf{P}re-training for \textbf{I}n-\textbf{C}ontext \textbf{L}earning), a framework to enhance the language models' in-context learning ability by pre-training the model on a large collection of ``intrinsic tasks'' in the general plain-text corpus using the simple language modeling objective. PICL encourages the model to infer and perform tasks by conditioning on the contexts while maintaining task generalization of pre-trained models. We evaluate the in-context learning performance of the model trained with PICL on seven widely-used text classification datasets and the \textsc{Super-NaturalInstrctions} benchmark, which contains 100+ NLP tasks formulated to text generation. Our experiments show that PICL is more effective and task-generalizable than a range of baselines, outperforming larger language models with nearly 4x parameters. The code is publicly available at \url{https://github.com/thu-coai/PICL}.
%In this work, we study enhancing language models' in-context learning ability in the pre-training stage. 
%We first notice that the pre-training corpus is not fully utilized and can be reorganized for language models to better learn to learn in the context.
%Specifically, we assume each paragraph in the pre-training corpus includes an intrinsic task. We design a retrieval-based method to gather paragraphs representing the same tasks and concatenate them as in-context learning examples. Pre-trained on these examples with a language modeling objective, the model learns to infer the implicit tasks and do in-context learning conditioning on the previous demonstrations.

%while obtaining better generalization on a diversity of downstream tasks.
%We also conduct extensive experiments to analyze the key factors of PICL and investigate how to pre-train language models for better in-context learning.

% 数据里面有 intrinsic tasks，并且用这些 task 去 train 有用

\end{abstract}

\section{Introduction}

{\let\thefootnote\relax\footnotetext{
$^\dagger$ Corresponding author.
}}

{\let\thefootnote\relax\footnotetext{
$^*$ Contribution during internship at Microsoft Research.
}}

Pre-trained language models (PLMs;~\citealp{plmsurvey, plm_qiu}) have shown strong abilities of learning and performing unseen tasks conditioning on several task examples or instructions in its context, which is called \textit{in-context learning} (ICL;~\citealp{gpt3}).
%~\footnote{We take a broad definition of ICL, including both learning from examples and human instructions, while previous works most refer ICL to learning from examples.}.
Compared to conventional fine-tuning methods, ICL adapts PLMs to downstream tasks only through inference, without parameter updates, which is computationally cheaper in practice and is closer to general AI.

However, PLMs trained on massive corpora to predict the next word given previous words are not explicitly taught to learn in the context. This makes ICL a surprising emergent ability but also indicates that the ICL ability of PLMs is not fully exploited. \citet{icl-function-class} has shown that by directly training to do ICL in a meta-learning paradigm, models show strong performance on learning simple function classes in the context. In practical NLP scenarios, previous works~\cite{metaicl, meta-in-context-tuning} also enhance the ICL performance by meta-fine-tuning PLMs on a large collection of downstream tasks and evaluating them on unseen tasks. However, the low diversity of human-annotated downstream tasks restricts the performance of the meta-tuned model. Direct training on downstream tasks also brings undesired bias on specific input formats, label spaces, or domains, which hurts the generalization of PLMs. 

\begin{figure*}[t]
    \centering
    \includegraphics[width=\linewidth]{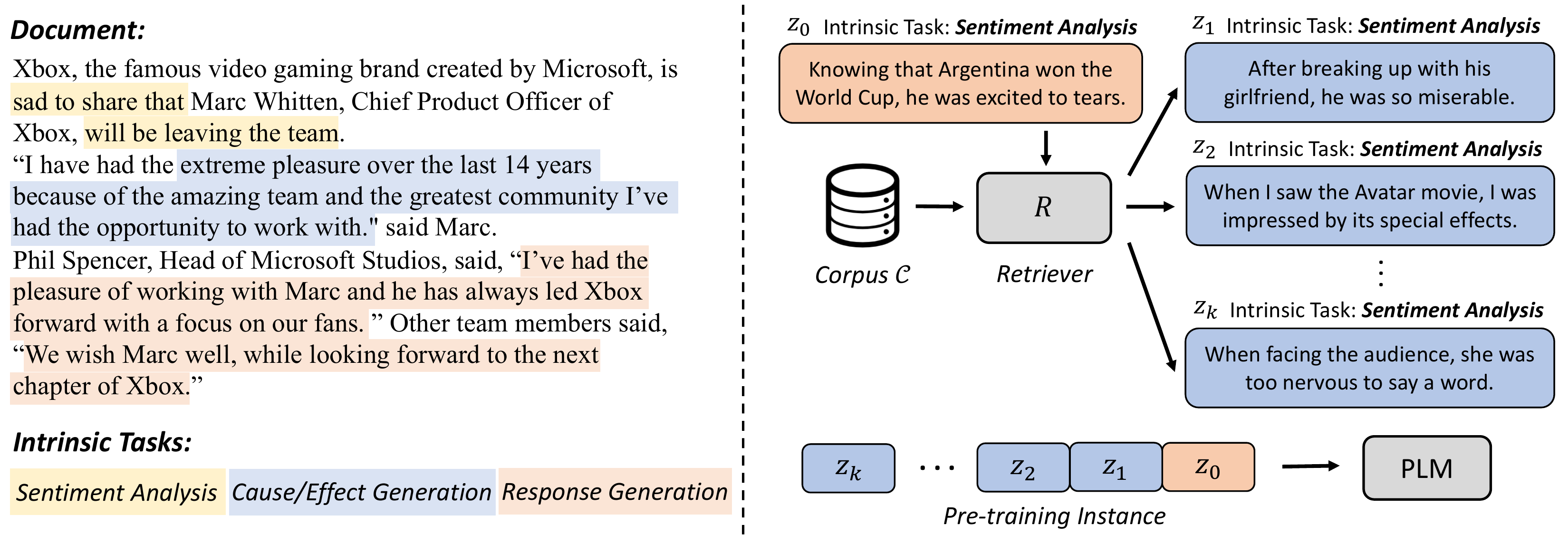}
    \caption{\textbf{Left:} An example of intrinsic tasks found in a document from the OpenWebText~\cite{openwebtext} corpus. \textbf{Right:} The overall framework of PICL. For each paragraph $z_0$ in the corpus $\mathcal{C}$, we retrieve $k$ paragraphs that share the same intrinsic task (Sentiment Analysis) as demonstrations and then concatenate them with $z_0$ to construct a pre-training instance. We compute the language modeling loss on the whole instance to train the model.}
    \label{fig:method}
    \vspace{-0.2em}
\end{figure*}

To enhance the ICL ability while maintaining generalization, we propose PICL (\textbf{P}re-training for \textbf{I}n-\textbf{C}ontext \textbf{L}earning), a framework that exploits the PLM's ICL ability by pre-training models on data automatically constructed from the general plain-text corpus. Our framework is based on a simple observation that many paragraphs in the text documents contain ``intrinsic tasks''. As shown in the left part of Figure \ref{fig:method}, each paragraph in the document contains an intrinsic task. When doing language modeling on each paragraph, models implicitly perform the corresponding intrinsic tasks simultaneously. This shares a similar idea with the prompt-learning paradigm~\cite{prompt_survey}, where downstream data examples from NLP tasks are transformed into text sequences, and the model learns to perform the original tasks when trained on the text sequences with language modeling. Different from the downstream data, text paragraphs contain more diverse intrinsic tasks and have little bias on input formats, label spaces, or domains because they are free-form texts from the large-scale general corpus. By gathering and concatenating paragraphs with the same intrinsic tasks (right part of Figure \ref{fig:method}), we can construct a meta-training dataset to pre-train the model to perform the intrinsic task conditioning on paragraphs in the context, and thereby improve the ICL ability.

We adopt a retrieval-based approach to gather paragraphs sharing the same intrinsic tasks from a general corpus. We first train an encoder to represent each paragraph in a vector space where paragraphs with the same intrinsic task have close embeddings. The encoder is trained with contrastive learning~\cite{supervised-constrastive} on a collection of downstream datasets by taking examples from the same tasks as positive pairs and those from different tasks as negative pairs. 
Then, treating any paragraph in the corpus as a query, we retrieve the paragraphs with close representations to the query, namely, sharing the same intrinsic task with the query. Finally, we concatenate the query and the retrieved paragraphs to get a pre-training instance. Note that although we use downstream datasets, the model is trained on instances constructed from the general corpus, which ensures its generalization.
%%%这个模型未来或许可以让大模型本身来问，无监督或弱监督地关联到某个任务上。
%%%就是 给一个提问：这个数据related to which task，LLM来回答。

%By doing language modeling on the example, the model learns to perform the intrinsic task with the help of the previous examples in the context.

We evaluate the ICL performance of the model pre-trained with PICL on seven widely-used text classification datasets and \textsc{Super-NaturalInstructions}~\cite{super-natural-instructions}, a benchmark whose test split contains more than 100 tasks formulated into text generation. Empirical results show the effectiveness of PICL, enabling the model to reach or even outperform larger models with nearly 4x parameters. Besides, we find that the PICL-trained model is more generalizable on various tasks than previous meta-fine-tuning methods. We also conduct extensive experiments to analyze several key factors of PICL.

\section{Method}

We first present an overview of PICL and then describe the details in the following sections. As shown in the right part of Figure \ref{fig:method}, we construct the pre-training instances from a corpus $\mathcal{C}$ consisting of paragraphs split from full documents by ``$\backslash$n''. For each paragraph $z_0$ in $\mathcal{C}$, we first use a retriever $R$ to find $k$ paragraphs $\{z_1, z_2, \cdots, z_k\}$ sharing the same intrinsic task (Sentiment Analysis) with $z_0$. Then the retrieved paragraphs are treated as demonstrations and concatenated with $z_0$ to form a pre-training instance: $z_k\oplus z_{k-1} \oplus \cdots \oplus z_1 \oplus z_0$. Finally, we adopt a language modeling objective to pre-train the model on the constructed instances.

In this way, the pre-training stage can be regarded as a meta-training process, where the model learns to solve the intrinsic task in $z_0$ conditioning on its context $z_k\oplus z_{k-1} \oplus \cdots \oplus z_1$.  Since $\mathcal{C}$ is a large-scale general corpus, it contains a variety of intrinsic tasks and little domain bias, which ensures the generalization of the pre-trained model. 

%Although we pre-train our model in a \textit{Example Sequence} format, we argue that pre-training improves the basic ability to learn from the context, and can also benefit the setting when the model receives a more free-form context.

\subsection{Retriever}
\label{sec:encoder}

% 现在这里的逻辑是按照 DPR 来的，结果和他们差不多（n-batch negative。没有在 method 里面写，作为细节放到后面的 setup 里面去了）

\begin{figure}[t]
    \centering
    \includegraphics[width=\linewidth]{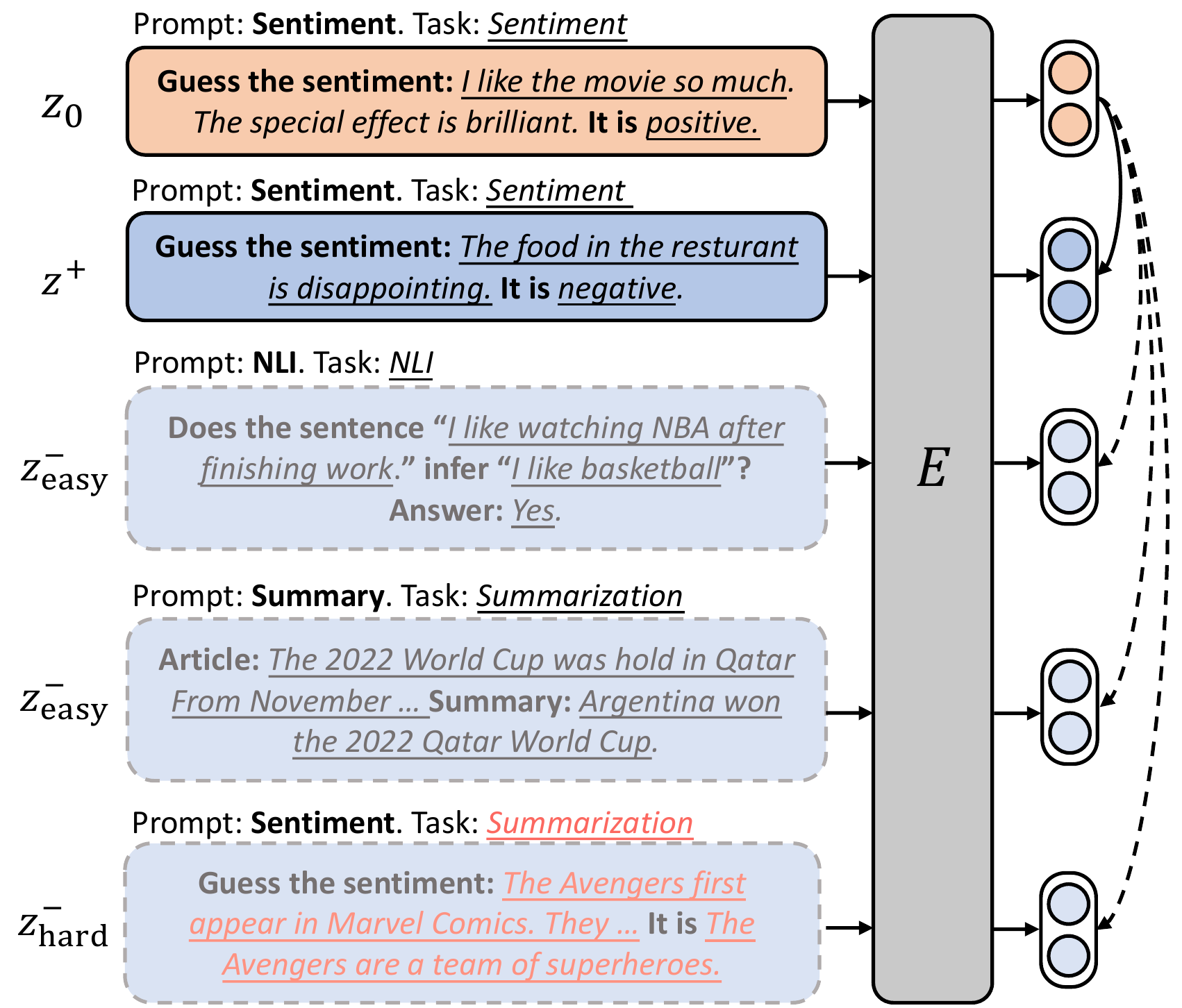}
    \caption{An example of how we construct the positive and negative pairs to train the task-semantics encoder $E$. The solid line means positive pairs, and the dashed lines mean negative pairs.}
    \label{fig:retriever}
    % \vspace{-0.5em}
\end{figure}

The main component of the retriever $R$ is a task-semantics encoder $E$ that represents a text paragraph as a $d$-dimensional vector in a space $V$, where paragraphs with the same intrinsic tasks have similar representations. We define the similarity between two paragraphs $z_0$ and $z$ using the dot product of their representations: $E(z_0) \cdot E(z)$.
% \begin{equation}
%     \small
%     \begin{aligned}
%     \operatorname{sim}(z_0, z) = E(z_0) \cdot E(z).
%     \end{aligned}
% \end{equation}

\paragraph{Encoder} We use RoBERTa$_{\text{BASE}}$~\cite{roberta} as the base model of $E$. The output vector is computed by averaging the last-layer representation of each token in the input paragraph.

\paragraph{Retrieval} We approximate that paragraphs whose representations are close to each other in $V$ share the same intrinsic task. Therefore, for every paragraph $z_0$ in $\mathcal{C}$, $R$ searches for $k$ paragraphs with embeddings closest to $E(z_0)$:
\begin{equation}
\small
\setlength{\abovedisplayskip}{8pt}
\setlength{\abovedisplayshortskip}{8pt}
\setlength{\belowdisplayskip}{8pt}
\setlength{\belowdisplayshortskip}{8pt}
\begin{aligned}
    R(z_0) = \{z_k, z_{k-1}, \cdots z_1\} = \operatorname{top-k}_z (E(z_0) \cdot E(z)).
\end{aligned}
\end{equation}
We employ the FAISS library~\cite{faiss} for efficient searching.

% \subsubsection{Training}

%To retrieve paragraphs that share the same intrinsic task with the query, 
\paragraph{Contrastive Training} We adopt contrastive learning~\cite{supervised-constrastive,dpr} to train the task-semantics encoder $E$. As shown in Figure \ref{fig:retriever}, we take two paragraphs with the same intrinsic task as positive pairs and those from different tasks as negative pairs. However, the annotation of a paragraph's intrinsic task is usually unavailable. To this end, we use a collection of downstream NLP datasets from various tasks whose examples are converted into text sequences with human-written prompts to train $E$. In this way, treating each text sequence as a paragraph, we can regard the corresponding downstream task as the intrinsic task annotation. We assume that the instances from all downstream tasks form a dataset $\mathcal{D}$. For each $z_0 \in \mathcal{D}$, we have a positive instance $z^+$ sharing the same task with $z_0$ and a set $\mathcal{N}(z_0)$ consisting of negative instances with different tasks than $z_0$, the loss function takes the form:
\begin{equation}
    \small
    \begin{aligned}
        & \mathcal{L}(z_0, z^+, \mathcal{N}(z_0)) \\
        & = -\log \frac{e^{E(z_0) \cdot E(z^+)}}{e^{E(z_0) \cdot E(z^+)} + \sum\limits_{z^- \in \mathcal{N}(z_0)} e^{E(z_0) \cdot E(z^-)}}.
    \end{aligned}
\end{equation}

%Assume that test sequences from all tasks form a unified set $\mathcal{D}$, for each $z \in \mathcal{D}$, we denote $\tau_z$ as its (intrinsic) task. $E$ is then trained with a supervised contrastive learning objective~\cite{supervised-constrastive} to represent test sequences from the same (intrinsic) task with similar embeddings .

%and generalize it to arbitrary paragraphs and intrinsic tasks in $\mathcal{C}$. 

%\paragraph{Data}  To ensure that $E$ can generalize to arbitrary paragraphs and intrinsic tasks in $\mathcal{C}$, we collect data from various tasks and specify each example with multiple prompts. We use 37 tasks ranging from text classification to generation and 320 prompts from PromptSource~\cite{p3}. Details of the tasks and prompts can be found in Appendix \ref{sec:app_data_detail}. We take two text sequences from the same tasks as positive pairs and those from different tasks as negative pairs.

\paragraph{Positive and Negative Instances} For each $z_0 \in \mathcal{D}$, we randomly sample a positive instance $z^+$ belonging to the same task with $z_0$ from $\mathcal{D} \backslash \{z_0\}$. As shown in Figure \ref{fig:retriever}, $\mathcal{N}(z_0)$ contains two kinds of negative instances: (1) Easy Negatives $z^-_{\text{easy}}$ sampled from $\mathcal{D}$ and belonging to different tasks than $z_0$. (2) Hard Negatives $z^+_{\text{hard}}$ sharing the same prompt with $z_0$ but containing mismatched tasks. For instance, in Figure \ref{fig:retriever}, we apply the prompt from the sentiment task to the summarization task to create the hard negative instance $z^-_{\text{hard}}$. This prevents the model from hacking the contrastive objective using prompts like ``Guess the sentiment'' and learning a trivial pattern matching but forces the model to extract task semantics from the whole paragraph.
%We denote $\tau_{k^-_{\text{hard}}}$ as the task of the original example (summarization), rather than the prompt (sentiment).

% \paragraph{Training Loss} We adopt the in-batch negative trick~\cite{dpr} to efficiently compute the loss. For each instance $q$ in a mini-batch $\mathcal{B}\subset \mathcal{D}$, we sample a positive instance $k^+$ from $\mathcal{D}$ that ensures $\tau_q = \tau_{k^+}$ ($q\ne k^+$) and construct several hard instances $k^-_{\text{hard}}$ that share the same prompt with $q$ but $\tau_q \ne \tau_{k^-_{\text{hard}}}$. We add all $k^+$ to $\mathcal{B}$ and put $k^-_{\text{hard}}$ in another batch $\mathcal{B}_{\text{Hard}}$. The training loss takes the following form:
% \begin{equation}
% \small
% \begin{aligned}
%     \mathcal{L}_E &= -\sum_{q\in \mathcal{B}} \frac{1}{|\mathcal{P}(q)|} \sum_{k \in \mathcal{P}(q)} \log \frac{e^{E(q)^\top E(k)}}{\sum_{k'\in \mathcal{A}(q)} e^{E(q)^\top E(k')}}, \\
% \end{aligned}
% \end{equation}
% where $\mathcal{P}(q) = \{k\in\mathcal{B}\backslash \{q\}:\tau_k=\tau_q\}$, $\mathcal{A}(q)=\mathcal{B}\cup\mathcal{B}_{\text{Hard}}\backslash \{q\}$  and $E(\cdot) \in V$.

%In Figure \ref{fig:method} the selected paragraphs share an intrinsic task ``\textit{Inferring the sentiment from the context}''. Then, we concatenate the retrieved and the query paragraph with a deliminator (``$\backslash$n'' in Figure \ref{fig:method}) to form the a pre-training example: $s=z_1\oplus z_2 \oplus, \cdots, \oplus z_k \oplus z_0$. Note that we always ensure $z_0$ appears in the end of $s$.

\subsection{Data Construction}
\label{sec:filter}
For each paragraph $z_0 \in \mathcal{C}$, we concatenate the retrieved paragraphs $\{z_1, z_2, \cdots, z_k\} = R(z_0)$ with $z_0$ to get a pre-training instance $z_k\oplus z_{k-1} \oplus \cdots \oplus z_1 \oplus z_0$. To improve the quality of the constructed data, we derive an approach to filter out instances that are less informative to ICL. We consider the following score to measure the informativeness of an instance based on the perplexity difference of the paragraphs in the instance before and after they are concatenated as a sequence:
\begin{equation}
    \small
    \begin{aligned}
        s = \frac{-\sum_{i=0}^k\log P(z_i) + \log P(z_k\oplus z_{k-1} \oplus \cdots \oplus z_0)}{|z_k\oplus z_{k-1} \oplus \cdots \oplus z_0|},
    \end{aligned}
    \label{eq:filter}
\end{equation}
where $|\cdot|$ is the length of a sequence and $P(\cdot)$ is the language modeling probability based on any uni-direct PLMs. Given a manually set threshold $\delta$, we retain the instances that satisfy $s > \delta$. This criterion leverages the original ICL ability of the PLM. If concatenating the paragraphs results in lower perplexity, they are more correlated and may be more informative for ICL. We finally construct a pre-training corpus containing $N$ instances $\mathcal{C}_{\text{pre-train}} = \{z^i_k\oplus z^i_{k-1} \oplus, \cdots, \oplus z^i_1\oplus z^i_0\}_{i=1}^N$.

\subsection{Pre-Training}
\label{sec:pre-train}
We pre-train the model with auto-regressive language modeling on $\mathcal{C}_\text{pre-train}$.
Unlike previous works~\cite{metaicl,meta-in-context-tuning}, which only compute the language modeling loss on the label tokens, we compute the loss on the whole sequence. There are two reasons for this choice. First, the intrinsic tasks are already in the natural language format, and it is unnecessary to split the input and the label. Second, we argue that computing loss on the whole sequence ensures a large token number in a forward batch, which is critical to maintaining the basic in-weights ability~\cite{data-distribution-icl}. Therefore, the loss function is:
\begin{equation}
\small
\setlength{\abovedisplayskip}{8pt}
\setlength{\abovedisplayshortskip}{8pt}
\setlength{\belowdisplayskip}{8pt}
\setlength{\belowdisplayshortskip}{8pt}
\begin{aligned}
    \mathcal{L}_{\text{ICL}}(\theta) = -\frac{1}{N} \sum_{i=1}^N \log P(z^i_k\oplus z^i_{k-1} \oplus \cdots \oplus z^i_0;\theta),
\end{aligned}
\end{equation}
% \subsubsection{Filtering}
% \paragraph{Searching Distance} An direct way for filtering is to drop the data with high searching distance. In practice, if an instance $z_1\oplus z_2 \oplus, \cdots, \oplus z_k \oplus z_0$ satisfies the following criterion, it is filtered out:
% \begin{equation}
%     \small
%     \begin{aligned}
%         \frac{\sum_{i=1}^k \operatorname{d}(E(z_i), E(z_0))}{k} > \delta_1,
%     \end{aligned}
% \end{equation}
% where $\delta_1$ is a threshold chosen manually.
where $\theta$ is the parameters of the model. In addition, we find that adding a language modeling loss $\mathcal{L_{\text{LM}}}(\theta)$ on the original full documents before being split into paragraphs benefits the performance. Therefore, the final optimization objective is:
\begin{equation}
    \small
    \setlength{\abovedisplayskip}{8pt}
\setlength{\abovedisplayshortskip}{8pt}
\setlength{\belowdisplayskip}{8pt}
\setlength{\belowdisplayshortskip}{8pt}
    \begin{aligned}
        \min_\theta \alpha \mathcal{L}_{\text{ICL}}(\theta) + (1-\alpha) \mathcal{L}_{\text{LM}}(\theta).
    \end{aligned}
\end{equation}
where we set $\alpha=0.5$ in our main experiments.

%We also find that mix-in a  When pre-trained on $s$ with $\mathcal{L}_\text{Full}(\theta)$, the model takes the intrinsic tasks as the meta-training tasks and learns to learn in the context. The diversity of intrinsic tasks and the little bias of $\mathcal{C}$ ensures the generalization of the pre-trained model. Although we pre-train our model in a \textit{Example Sequence} format, we argue that pre-training improves the basic ability to learn from the context, and can also benefit the setting when the model receives a more free-form context.

\section{Experimental Setup}

\subsection{Pre-training Data}
We merge \textsc{OpenWebText}~\cite{openwebtext}, \textsc{WikiCorpus}~\cite{wikidump}, and \textsc{BookCorpus}~\cite{bookcorpus} to construct the pre-training data, where full documents are split into paragraphs by ``$\backslash$n''. The corpus $\mathcal{C}$ consists of 80M paragraphs, totaling about 30GB.
% \footnote{80M is the largest corpus size that all the paragraph embeddings can fit into the GPU memory for efficient search.}
For each paragraph, we search for $k=20$ demonstrations and concatenate them until 1024 tokens, the maximum input length constraint of the language model we used. This ensures that the model sees various demonstration numbers during pre-training. We use GPT2-Large~\cite{gpt2} to compute $P(\cdot)$ in Equation ~\ref{eq:filter} and set $\delta = 0.0$ for filtering. More details of data processing and statistics are shown in Appendix \ref{sec:app_pretrain_corpus}.

\subsection{Baselines}
We consider four baselines in our experiments:

\begin{itemize}[leftmargin=12pt,noitemsep,topsep=5pt]
    % \vspace{-5pt}
    \item \textbf{VanillaICL} directly prompts a PLM with the concatenation of training examples to do ICL. 
    \item \textbf{ExtraLM} further pre-trains the PLM on the original full documents before being split into paragraphs with the language modeling objective.
    \item \textbf{Self-Sup}~\cite{self-sup} designs four self-supervised pre-training objectives, including Next Sentence Generation, Masked Word Prediction, Last Phrase Prediction, and Classification, to enhance the ICL performance. We conduct the self-supervised pre-training on our merged corpus for a fair comparison.
    \item \textbf{MetaICL}~\cite{metaicl} meta-trains the model on a large collection of downstream human-annotated datasets for learning to learn in context. The meta-training instances are constructed by concatenating several training examples in each dataset to a single text sequence. We replicate the method on the training set of our task-semantics encoder for a fair comparison. 
    % \vspace{-8pt}
\end{itemize}

\subsection{Evaluation}
We evaluate the model trained with PICL on two kinds of downstream tasks. 
\paragraph{Few-Shot Text Classification} We consider seven widely-used text classification datasets, including SST-2~\cite{sst-2}, SST-5~\cite{sst-2}, Subj~\cite{subj}, MR~\cite{mr}, RTE~\cite{rte}, CB~\cite{cb}, and AGNews~\cite{agnews} to evaluate the few-shot ICL performance of the trained models (see Appendix \ref{sec:app_eval_data_detail} for more details). Note that these tasks are not included in the training set of the task-semantics encoder. We randomly sample 4 or 8 demonstrations from the official training sets of each dataset. Effects of other demonstration numbers can be found in Section \ref{sec:exp_demo}. We compute the average accuracy scores on at most 1000 samples from the validation split of each dataset across five random seeds for selecting demonstrations.

\paragraph{Instruction Following} To test the generalization of PICL, we also evaluate the trained model on a larger range of tasks with more free-form inputs, including both human instructions and few-shot examples. We use the test split of \textsc{Super-NaturalInstructions}~\cite{super-natural-instructions} as the benchmark and exclude the tasks that appear in the training set of the task-semantics encoder, resulting in 105 evaluation tasks (see Appendix \ref{sec:app_sni} for a full list of tasks). Each task is specified with a human-written instruction and two or three demonstrations. We follow \citet{super-natural-instructions} to formulate all tasks to the text generation format and score the outputs with ROUGE-L~\cite{rouge}.

%We build the context with the \textit{Sample Sequence} format, which is commonly adopted in previous works~\cite{gpt3}. We use datasets from text classification, text generation, and multiple-choice tasks to evaluate our model. Detailed dataset information and the templates we use can be found in Appendix ***.  

\begin{table*}[t]
\small
\centering
\begin{tabular}{lllccccccc|c}
\toprule
 % \multirow{2}{*}{Method}            & \multirow{2}{*}{Param.} & \multicolumn{8}{c}{\textit{Text Classification}} \\
% \cmidrule(lr){3-10}
Shot & Method                              & Param. & SST2              & SUBJ              & MR                    & RTE                   & AgNews                    & CB                    & SST5          & Average\\ \midrule
\multirow{7}{*}{4-shot}
& VanillaICL                          & 770M & 67.5$_{9.2}$         & 57.7$_{7.8}$          & 50.3$_{0.3}$          & 50.8$_{1.7}$          & 67.5$_{2.3}$              & 68.1$_{2.4}$          & 24.4$_{5.4}$ & 55.2$_{0.5}$\\
& \textcolor{gray}{VanillaICL} & \textcolor{gray}{1.5B} & \textcolor{gray}{74.9$_{9.7}$} & \textcolor{gray}{65.2$_{10.0}$} & \textcolor{gray}{61.9$_{6.5}$} & \textcolor{gray}{50.4$_{0.4}$} & \textcolor{gray}{65.6$_{4.8}$} & \textcolor{gray}{67.8$_{5.6}$} & \textcolor{gray}{32.4$_{4.6}$} & \textcolor{gray}{59.7$_{2.5}$}\\ 
& \textcolor{gray}{VanillaICL} & \textcolor{gray}{2.7B} & \textcolor{gray}{75.0$_{7.5}$} & \textcolor{gray}{65.4$_{2.9}$} & \textcolor{gray}{71.4$_{13.3}$} & \textcolor{gray}{49.8$_{1.8}$} & \textcolor{gray}{65.6$_{2.8}$} & \textcolor{gray}{60.0$_{2.1}$} & \textcolor{gray}{32.1$_{5.4}$} & \textcolor{gray}{59.9$_{1.1}$}\\ 
\cmidrule(l){2-11}
& ExtraLM                             & 770M & 68.9$_{11.3}$        & 63.9$_{6.4}$          & 60.3$_{6.4}$          & 51.2$_{1.7}$          & 64.5$_{1.5}$              & 63.7$_{5.3}$          & 27.8$_{5.1}$ & 57.2$_{2.1}$\\
% Random                              & 770M & 73.6$_{11.8}$        & 56.9$_{6.4}$          & \underline{73.0$_{9.0}$}      & 50.8$_{0.7}$      & 59.3$_{3.3}$      & 67.6$_{3.4}$      & 28.4$_{3.8}$ & 58.5$_{2.4}$\\
& Self-Sup                            & 770M & 55.0$_{7.4}$         & 50.3$_{0.6}$          & 59.7$_{3.5}$          & 52.2$_{2.0}$          & 50.3$_{7.0}$              & 63.4$_{7.1}$          &  28.8$_{3.3}$ & 51.4$_{2.2}$ \\ 
& MetaICL                             & 770M & 69.8$_{4.0}$         & 63.5$_{4.6}$          & 65.6$_{7.5}$          & \textbf{57.6$_{2.3}$} & 66.3$_{2.4}$              & 65.2$_{3.0}$          &  31.7$_{2.1}$ & 60.0$_{1.5}$ \\ 
\cmidrule(l){2-11}
& PICL                                & 770M & \textbf{79.7$_{8.6}$} & \textbf{66.8$_{7.4}$} & \textbf{81.0$_{1.3}$} & 54.5$_{1.8}$         & \textbf{67.7$_{3.4}$}     & \textbf{69.6$_{4.3}$}          & \textbf{34.8$_{4.0}$} & \textbf{64.4$_{1.6}$}\\
\midrule
\multirow{7}{*}{8-shot}
& VanillaICL                          & 770M & 68.7$_{6.0}$         & 66.6$_{9.8}$          & 60.2$_{5.5}$          & 51.8$_{1.6}$          & 60.2$_{5.6}$              & 68.8$_{3.2}$          & 31.4$_{3.8}$ & 58.2$_{2.9}$\\
& \textcolor{gray}{VanillaICL} & \textcolor{gray}{1.5B} & \textcolor{gray}{72.1$_{12.6}$} & \textcolor{gray}{63.4$_{6.5}$} & \textcolor{gray}{63.3$_{5.4}$} & \textcolor{gray}{52.7$_{2.8}$} & \textcolor{gray}{54.2$_{8.4}$} & \textcolor{gray}{\textbf{70.4$_{5.7}$}} & \textcolor{gray}{33.5$_{3.3}$} & \textcolor{gray}{58.6$_{2.5}$}\\ 
& \textcolor{gray}{VanillaICL} & \textcolor{gray}{2.7B} & \textcolor{gray}{71.0$_{11.6}$} & \textcolor{gray}{65.2$_{4.0}$} & \textcolor{gray}{70.4$_{6.3}$} & \textcolor{gray}{51.3$_{2.0}$} & \textcolor{gray}{63.1$_{2.4}$} & \textcolor{gray}{69.6$_{4.0}$} & \textcolor{gray}{\textbf{34.1$_{2.8}$}} & \textcolor{gray}{60.6$_{3.2}$}\\
\cmidrule(l){2-11}
& ExtraLM                             & 770M & 69.7$_{3.4}$         & 65.2$_{6.5}$          & 63.6$_{6.0}$          & 52.6$_{1.6}$          & 58.9$_{7.0}$              & 69.6$_{3.8}$          & 32.2$_{4.7}$ & 58.8$_{1.6}$\\
% Random                              & 770M & 73.6$_{11.8}$        & 56.9$_{6.4}$          & \underline{73.0$_{9.0}$}      & 50.8$_{0.7}$      & 59.3$_{3.3}$      & 67.6$_{3.4}$      & 28.4$_{3.8}$ & 58.5$_{2.4}$\\
& Self-Sup                            & 770M & 61.4$_{6.5}$         & 54.3$_{4.5}$          & 73.8$_{8.1}$          & 53.0$_{2.4}$          & 52.1$_{3.8}$              & 63.0$_{6.9}$          &  33.7$_{1.8}$ & 55.9$_{2.1}$ \\ 
& MetaICL                             & 770M & 73.6$_{6.2}$         & 67.2$_{8.8}$          & 70.1$_{5.6}$          & \textbf{53.6$_{2.1}$} & 56.1$_{0.7}$              & 65.8$_{4.1}$          &  33.7$_{4.7}$ & 60.0$_{2.2}$ \\ 
\cmidrule(l){2-11}
& PICL                                & 770M & \textbf{78.0$_{10.6}$} & \textbf{69.3$_{9.5}$} & \textbf{77.5$_{5.0}$} & 53.0$_{1.6}$        & \textbf{64.7$_{4.4}$}     & \textbf{70.4$_{2.1}$} & \textbf{34.1$_{3.8}$} & \textbf{63.9$_{1.3}$}\\

\bottomrule
\end{tabular}
\caption{Main results of few-shot text classification. We report the average accuracy scores and the standard deviations across 5 random seeds for selecting demonstrations. We use GPT2-Large (770M), GPT2-xLarge (1.5B), and GPT-\textit{Neo} (2.7B) for VanillaICL. The best scores on each dataset under 4 or 8 evaluation shots are in \textbf{boldface}.}
\label{tab:icl}
% \vspace{-0.5em}
\end{table*}

\subsection{Settings}
\paragraph{Retriever} We train the task-semantics encoder on 37 tasks (see Appendix \ref{sec:app_data_detail}) using up to 10K examples per task. To enhance generalization, we apply multiple prompts from PromptSource~\cite{p3} to one example and use 320 prompts in all. We use the in-batch negative trick~\cite{simclr} to compute the contrastive loss. We set the learning rate to $5\times10^{-5}$, the batch size to 64, and construct 4 hard negatives for each instance. The encoder is trained from RoBERTa$_{\text{Base}}$ for 1 epoch.

\paragraph{Language Model} We test PICL based on the 770M GPT2-Large~\cite{gpt2} unless otherwise specified. Results on larger models can be found in Appendix \ref{app:large_model}. To save computational resources, we train the model from its pre-trained checkpoints. We also test the VanillaICL performance of larger models, including GPT2-xLarge~\cite{gpt2} (1.5B) and GPT-\textit{Neo}~\cite{gpt-neo} (2.7B) for reference.

\paragraph{Pre-Training} We set the maximum learning rate to $1\times 10^{-6}$ and use the “inverse square root” scheduler~\cite{transformer} with 1000 steps warmup. The model sees 131K tokens in a step and is pre-trained for 100K steps. It takes less than a day to finish pre-training on 64 V100 32G GPUs.

\section{Results}
% \subsection{Main Results}
\subsection{Few-Shot Text Classification}
\label{sec:exp_main}

Table \ref{tab:icl} shows the results of few-shot text classification, from which we have 3 observations.

\textit{First}, among the baselines with 770M parameters, simply further training the model on our corpus with language modeling improves the performance (ExtraLM). This is likely due to the higher domain diversity of our corpus. MetaICL is helpful on most datasets, which verifies the effectiveness of meta-training for ICL. Self-Sup fails to bring benefits on most datasets against VanillaICL, probably because the constrained label space of the Classification training task (only contains ``True'' and ``False'') brings bias to the model's output. This emphasizes the importance of using training objectives with little bias.

\begin{figure*}
    \centering
    \includegraphics[width=\linewidth]{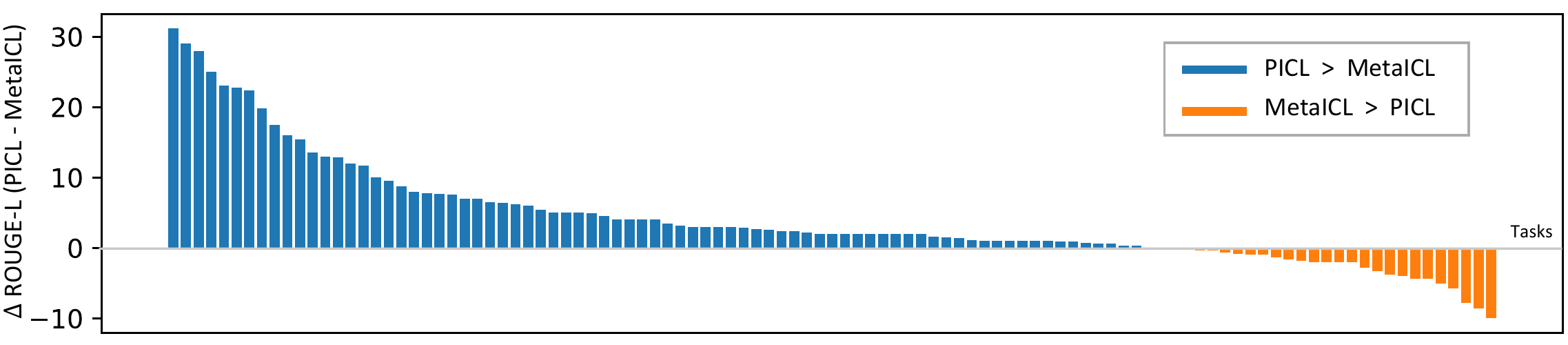}
    \caption{Comparison between PICL and MetaICL on \textsc{Super-NaturalInstructions}~\cite{super-natural-instructions}. Each bar represents an evaluation task. The y-axis means the ROUGE-L score difference between the two methods.}
    \label{fig:picl_vs_metaicl}
\end{figure*}

\textit{Second}, we observe that the PICL-trained model outperforms the baselines with the same model sizes by a large margin on most datasets across different shots, verifying the effectiveness of PICL. An exception is RTE, where MetaICL performs the best. We speculate the reason is that some training tasks of MetaICL share the same label space with RTE (``Yes''/``No''), such as paraphrase identification. \citet{rethinking-demons} has shown that the label space plays a vital role in ICL, which explains the good performance of MetaICL on RTE. 

\textit{Thrid}, comparing models across different sizes, we find that increasing the model parameters boosts the performance, but PICL enables the 770M model to beat a 2.7B counterpart. This indicates that the ICL ability can be enhanced not only through scaling up the parameters. Improving the structure of the pre-training data is also beneficial. In Appendix \ref{app:large_model}, we can see that PICL is also effective when applied to a 1.5B model.

\subsection{Instruction Following}

\begin{table}[t]
    \centering
    \small
    \begin{tabular}{llc}
    \toprule
      Model         & Param. & ROUGE-L \\
      \midrule
      VanillaICL   & 770M &   34.3  \\
      \textcolor{gray}{VanillaICL}  & \textcolor{gray}{1.5B} &   \textcolor{gray}{34.9}  \\
      \textcolor{gray}{VanillaICL} & \textcolor{gray}{2.7B} & \textcolor{gray}{37.3} \\
    \midrule
      ExtraLM       & 770M &   34.6 \\
      Self-Sup      & 770M &   30.5 \\
      MetaICL       & 770M &   35.3 \\
    \midrule
      PICL          & 770M &   \textbf{37.6} \\
    \bottomrule
    \end{tabular}
    \caption{Results of instruction following on \textsc{Super-NaturalInstructions}. We report the average ROUGE-L score across all 105 evaluation tasks.}
    \label{tab:super_ni}
    % \vspace{-1em}
\end{table}

The results on \textsc{Super-NaturalInstructions} are shown in Table \ref{tab:super_ni}. We can see that PICL achieves higher overall instruction following performance than the baselines, outperforming a larger model with about 4x parameters.

In Figure \ref{fig:picl_vs_metaicl}, we compare the per-task performance of PICL and MetaICL because they share the most similar setting where human-annotated downstream datasets are used. We observe that PICL outperforms MetaICL on about 3/4 of evaluation tasks, indicating that compared to fine-tuning directly on downstream tasks, pre-training on intrinsic tasks constructed from the general plain-text corpus brings better ICL ability and ensures higher generalization performance across a broad range of tasks (see Appendix \ref{app:res_nat} for more details). 

Most tasks where MetaICL beats PICL belong to text classification whose output spaces are ``Yes/No'' or ``True/False''. This matches the second observation in Section \ref{sec:exp_main}, where MetaICL predicts ``Yes/No'' well because of training on tasks that share the same label spaces. On the other hand, PICL performs much better on text generation, or tasks whose output spaces share the same semantics with ``Yes/No'' but use label words not in the training tasks of MetaICL (e.g., ``Correct/Wrong''). This indicates that direct training on downstream datasets causes overfitting to specific labels. There are also tasks where PICL performs similarly to MetaICL, such as reasoning and word analogy. We notice that the improvements of PICL and MetaICL on these tasks are also marginal against VanillaICL probably because these tasks rely more on the ``in-weights learning'' ability~\cite{data-distribution-icl}, rather than in-context learning.

% \begin{table*}[t]
% \small
% \centering
% \begin{tabular}{@{}lllllllllllll@{}}
% \toprule
%                & Coref. & Entail. & Rewrite. & CECLS & TitleGen. & Dial. & AnsCLS & D2T & Tag. & Analogy & Overlap & Overall \\\midrule
% Vanilla        &        &         &          &              &            &          &            &           &         &         &         & 34.3    \\
% LM             &        &         &          &              &            &          &            &           &         &         &         &         \\
% Random         &        &         &          &              &            &          &            &           &         &         &         &         \\
% Self-sup       &        &         &          &              &            &          &            &           &         &         &         &         \\
% MetaICL        &        &         &          &              &            &          &            &           &         &         &         &         \\
% PICL           &        &         &          &              &            &          &            &           &         &         &         &         \\
% \ \ + MetaICL           &        &         &          &              &            &          &            &           &         &         &         &         \\
% \bottomrule
% \end{tabular}
% \end{table*}

\subsection{Analysis}
\paragraph{Effect of Retriever}
We compare different approaches to retrieve paragraphs and test the final model performance. We try randomly selecting paragraphs (Random), retrieving using the non-parametric approach (BM25), encoding each paragraph with the original pre-trained encoder as it is (RoBERTa), or using the encoder for sentence similarity~\cite{sentence-bert} (SRoBERTa). We also study different numbers of hard negatives (0, 1, 4) and downstream tasks (7, 24, 37) to train the task-semantics encoder in PICL. From the results in Table \ref{tab:retriever}, we can see that all retrieval methods except Random bring improvements against VanillaICL on both text classification and instruction following settings, indicating that improving the coherence of the paragraphs in the pre-training data benefits ICL. Using the task-semantics encoder in PICL achieves the best performance, showing the importance of retrieving paragraphs based on task semantics rather than word overlap or sentence meanings. Comparing different settings to train the task-semantics encoder, we observe that increasing the number of hard negatives and training tasks improves the final performance. This is in line with previous works~\cite{dpr,simclr,moco} that more challenging hard negatives benefit contrastive learning.
% Besides, using 7 tasks still outperforms the 2.7B GPT-\textit{Neo}, which means that we do not 

% \begin{table}[t]
%     \centering
%     \small
%     \begin{tabular}{lcc}
%     \toprule
%      \multirow{2}{*}{Retriever} &  Classification & \textsc{Sup-NatInst.}   \\
%                                 & Accuracy & ROUGE-L       \\
%      \midrule
%      Random         &     &                             \\ 
%      BM25           &     &                             \\
%      RoBERTa        &      &                               \\
%      SRoBERTa       &      &                               \\
%      \midrule
%      PICL Encoder (0)  &       &                        \\  
%      PICL Encoder (1)  &     &                               \\       
%      PICL Encoder (4)  &     &                               \\       
     
%     \bottomrule
%     \end{tabular}
%     \caption{Caption}
%     \label{tab:my_label}
% \end{table}

\begin{table}[t]
    \centering
    \small
    \begin{tabular}{@{}lrrcc@{}}
    \toprule
     \multirow{2}{*}{Retriever} & \multirow{2}{*}{$n_{\text{HardNeg.}}$}  & \multirow{2}{*}{$n_{\text{Tasks}}$}    &  CLS & \textsc{Sup-NI}         \\
                                &   &     & Accuracy & ROUGE-L        \\
     \midrule
     VanillaICL                        & -  & -  & 55.2 &  34.3       \\ \midrule
     % VanillaICL (2.7B)                 & -  & -  & 59.9 &  37.3       \\ \midrule
     Random                     & -  & -  & 56.7 &  29.3       \\ 
     BM25                       & -  & -  & 59.2 &  34.5       \\
     RoBERTa                    & -  & -  & 58.7 &  34.6       \\
     SRoBERTa                   & -  & -  & 59.0 &  35.0       \\
     \midrule
\multirow{5}{*}{PICL}           & 0  & 37 & 62.2 &  36.4       \\  
                                & 1  & 37 & 63.1 &  36.5        \\
                                & 4  &  7 & 61.6 &  35.4       \\
                                & 4  & 24 & 63.4 &  36.6       \\   
                                & 4  & 37 & \textbf{64.4} & \textbf{37.6}        \\
    \bottomrule
    \end{tabular}
    \caption{Comparison of different retrievers. $n_{\text{HardNeg.}}$ and $n_{\text{\text{Tasks}}}$ means the number of hard negatives and downstream tasks to train the task-semantics encoder in PICL. ``CLS Accuracy'' means the average accuracy scores on text classification tasks. ``\textsc{Sup-NI} ROUGE-L'' means the average ROUGE-L scores across the tasks in \textsc{Super-NaturalInstructions}.}
    \label{tab:retriever}
\end{table}

\paragraph{Effect of Demonstration Numbers}
\label{sec:exp_demo}
% \begin{table}[t]
%     \centering
%     \small
%     \begin{tabular}{|c|c|c|c|c|c|c|}
%     \hline
%                 &  0    & 2     & 4     & 8     & 16    & Avg.\\ \hline
%      ICL        &  53.0 & 50.9  & 55.2  & 60.5  & 60.0  & 55.9\\ \hline
%      MetaICL    &  53.1 & 53.8  & 60.0  & 59.4  & 61.7  & 57.6\\ \hline
%      PICL (2)   &  55.1 & 56.3  & 61.3  & 60.3  & 61.0  & 58.8\\  \hline
%      PICL (4)   &  55.3 & 56.8  & 62.5  & 61.8  & 61.9  & 60.0\\  \hline
%      PICL (8)   &  53.2 & 56.6  & 63.6  & 64.0  & 63.6  & 60.2\\  \hline
%      PICL (16)  &  52.6 & 54.7  & 62.6  & 63.1  & 62.4  & 59.1\\  \hline
%      PICL       &  53.6 & 57.0  & 64.4  & 63.7  & 63.8  & 60.5\\ \hline
%     \end{tabular}
%     \caption{Caption}
%     \label{tab:demo}
% \end{table}

\begin{figure}
    \centering
    \includegraphics[width=0.9\linewidth]{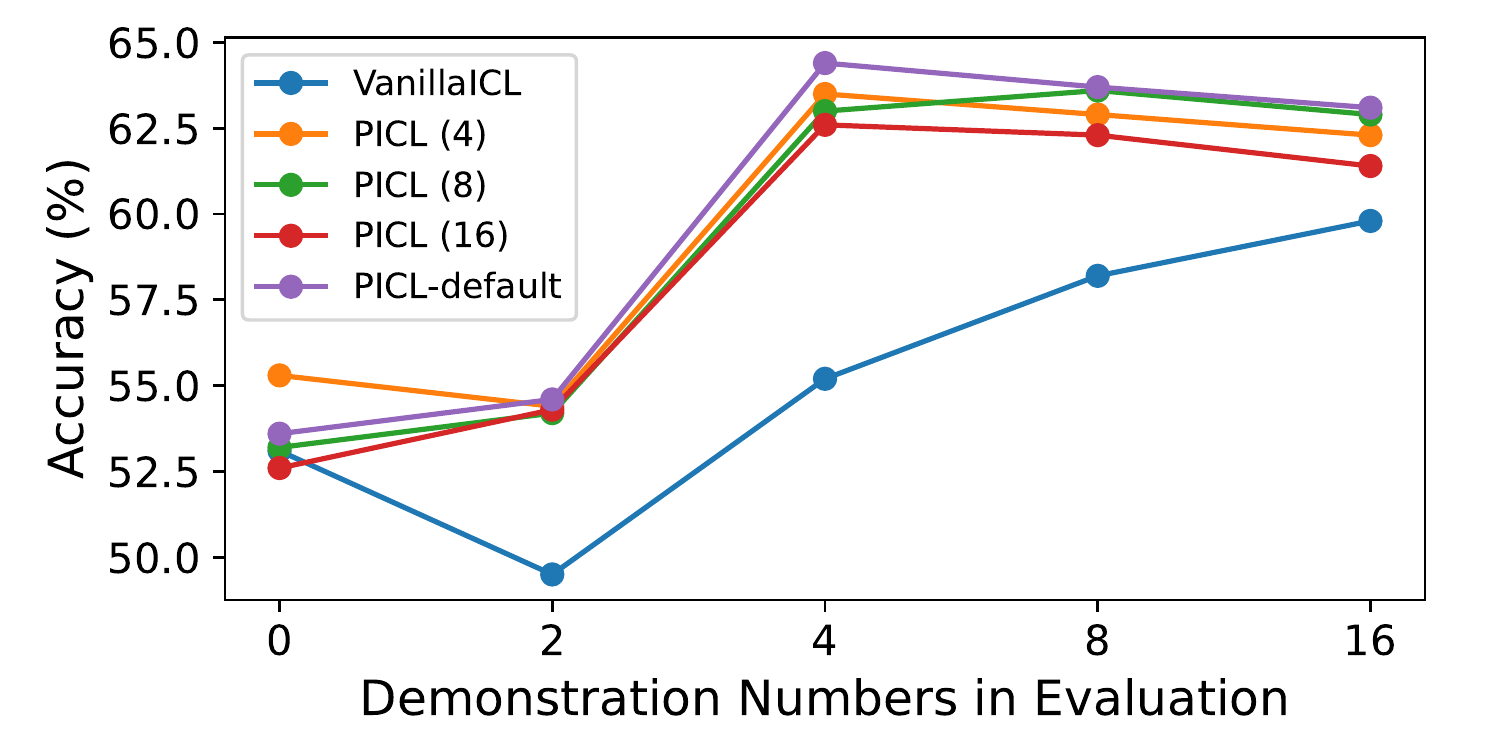}
    \caption{Average text classification accuracy when the pre-training instances contain different demonstration numbers in PICL (the number in the brackets). ``PICL-default'' means using a mixture of demonstration numbers as in previous experiments. }
    \label{fig:exp_demo}
\end{figure}

Training with PICL brings two benefits: (1) PLMs learn a format where demonstrations from the same task are concatenated as the prefix, which is beneficial when the model is evaluated under the same number of demonstrations. (2) PLMs learn a better ability to infer and perform tasks from the context, even when the demonstration numbers in evaluation and pre-training do not match. To differentiate these effects, we conduct pre-training on instances containing only 4, 8, or 16 demonstrations and test the trained models under different text classification shots. Results in Figure \ref{fig:exp_demo} show that when pre-trained with different demonstration numbers, the models generalize well to unseen demonstration numbers in evaluation, achieving similar performance with the default setup where the model sees various demonstration numbers in pre-training (PICL-default). This indicates that the models learn more than the input formats in PICL. 

%In addition, we find that training with various demonstration numbers performs relatively the best (PICL-default).

\begin{figure}[t]
    \subfigbottomskip=0cm
    \centering
    \subfigure[Effect of $\delta$] { \label{fig:filter} 
    \centering
    \includegraphics[width=0.9\linewidth]{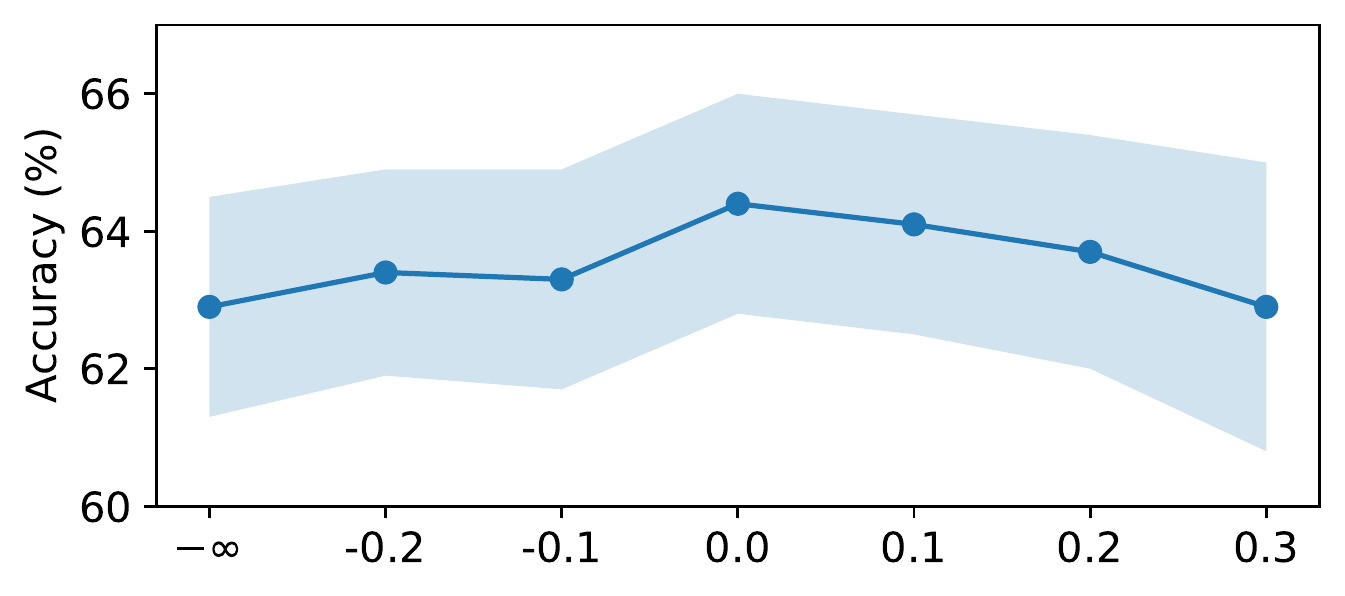} 
    } \\
    \subfigure[Effect of $\alpha$] { \label{fig:lm} 
    \centering
    \includegraphics[width=0.9\linewidth]{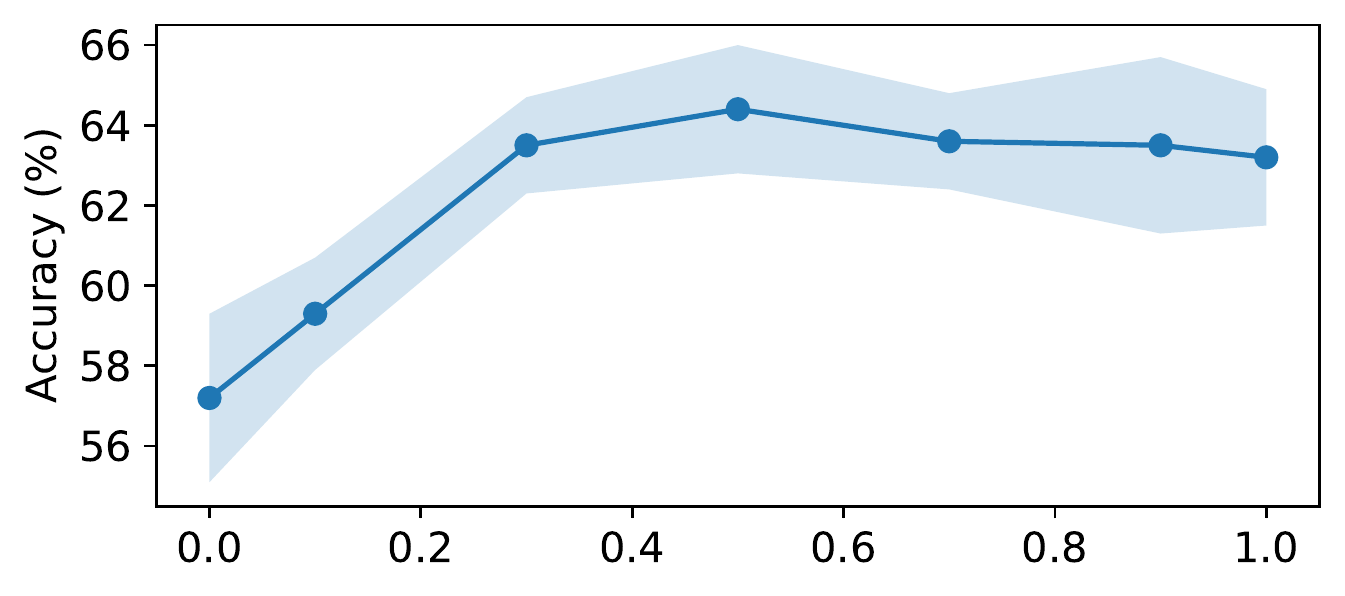} 
    }
    \caption{Hyper-parameter analysis. \textbf{(a)}: average 4-shot text classification accuracy as a function of $\delta$ for filtering. $-\infty$ means we do not conduct filtering. \textbf{(b)}: average 4-shot text classification accuracy as a function of $\alpha$ to control the proportion of the full-documents.} 
\end{figure} 

\begin{figure}[t]
    \subfigbottomskip=0cm
    \centering
    \subfigure[Effect of Data Amount] { \label{fig:data_amount} 
    \centering
    \includegraphics[width=0.9\linewidth]{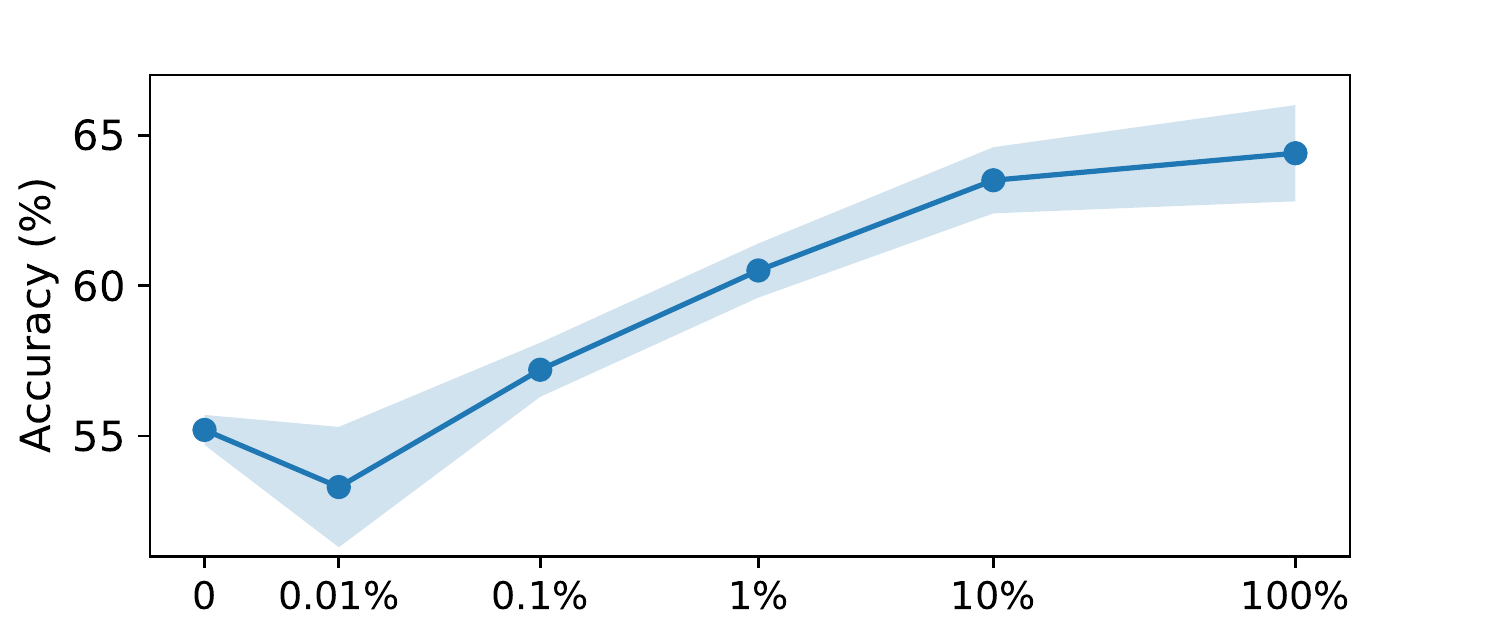} 
    } \\
    \subfigure[Data Comparison] { \label{fig:data_quality} 
    \centering
    \includegraphics[width=0.9\linewidth]{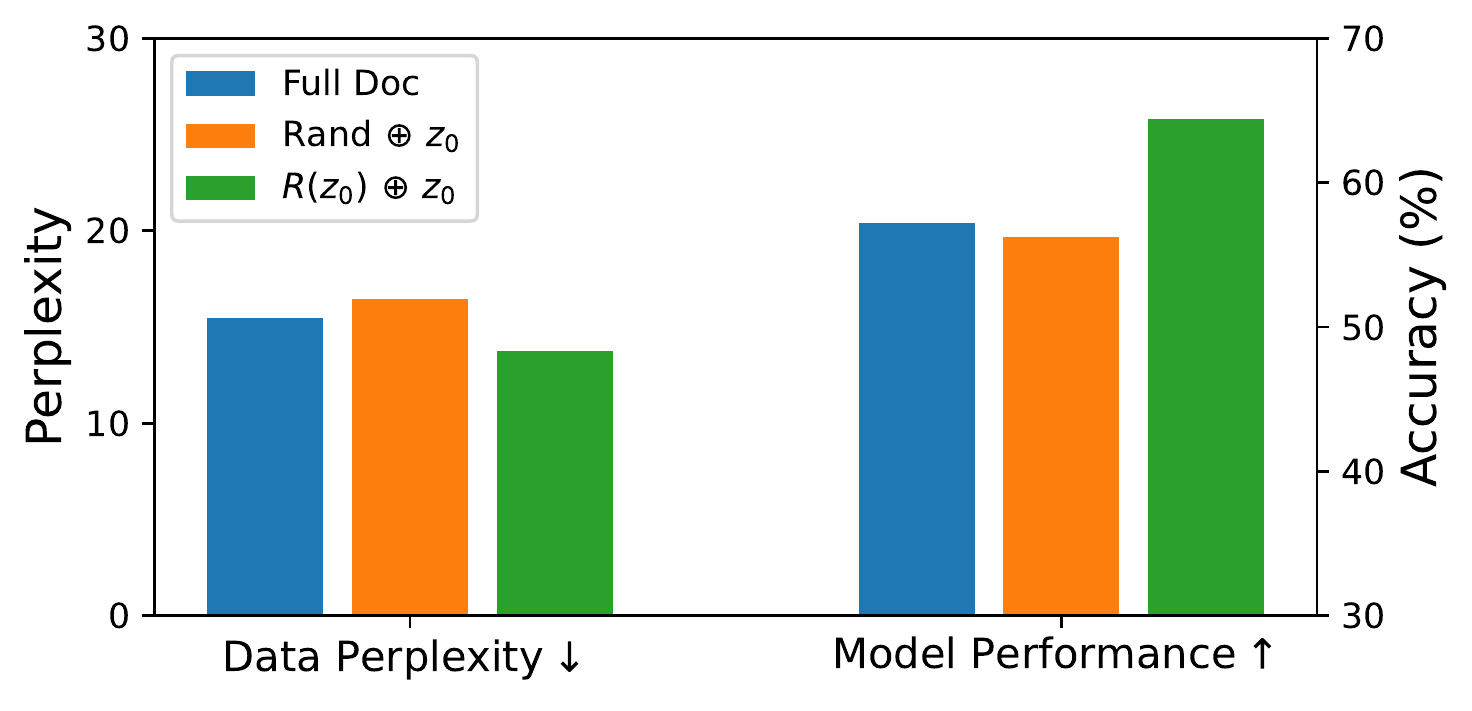} 
    }
    \caption{Data analysis. \textbf{(a)}: the average 4-shot text classification accuracies when constructing data using different proportions of the original corpus. \textbf{(b)}: perplexity of full-document data (Full Doc), random retrieved data (Rand $\oplus \ z_0$) and PICL data ($R(z_0) \oplus z_0$) based on GPT-J (6B) and the corresponding model performance.} 
\end{figure} 

\paragraph{Effect of Filtering}
We try different threshold values $\delta$ for filtering and report the scores on text classification tasks in Figure \ref{fig:filter}, while controlling the sizes of the constructed pre-training data the same. We find that $\delta=0$ yields the best performance, which means we retain an instance if and only if the perplexity of individual paragraphs is higher than that of the concatenated sequence (Equation \ref{eq:filter}). For lower $\delta$, the pre-training data contain too many uninformative instances for ICL. For larger $\delta$, we speculate that the filtering process relies on the original GPT2-Large too much. Since we also pre-train based on GPT2-Large, the filtering process reduces the signals in the constructed data beyond the base model's ability.

\paragraph{Effect of Full Documents}
In Figure \ref{fig:lm}, we report the model performance on text classification tasks when using different choices of $\alpha$, which controls the proportion of the full-document data. We find that balancing the constructed and full-document data performs the best ($\alpha=0.5$). When $\alpha$ is too large, the model is trained mostly on our constructed data and overfits its bias inevitably introduced by the task-semantics encoder in the data construction process. When $\alpha$ is too small, our method degenerates into ExtraLM.

\paragraph{Effect of Data Amount}
We study the size effect of the corpus used to construct the pre-training data in PICL and report the performance on text classification tasks in Figure \ref{fig:data_amount}. We conduct the data construction on 0.01\%, 0.1\%, 1\%, and 10\% of the original 80M paragraphs (100\%) and pre-train models for at most 100K steps until the validation losses begin to increase. From the results, we conclude that when the corpus is small, pre-training with the constructed data hurts the performance because the search library is too small to find paragraphs sharing the same intrinsic tasks. Training on small data for multiple epochs also causes overfitting. When the corpus contains more than 80K paragraphs (0.1\%), adding more data constantly improves the performance, which is consistent with the scaling law~\cite{scaling_law}. 

\paragraph{Data Comparison}
We compare the usefulness of different pre-training data to enhance the ICL ability. In addition to the final model performance, we borrow the thoughts for designing the filtering criterion in Section \ref{sec:filter} to measure the usefulness of a pre-training instance by computing the perplexity using a reference large PLM: GPT-J~\cite{gpt-j} with 6B parameters. Lower perplexity means the correlation within the instance is higher and is intuitively more helpful for enhancing the ICL ability. In Figure \ref{fig:data_quality}, we show the perplexity and the final model performance of 3 pre-training data: original full documents before being split into paragraphs (Full Doc), concatenation of randomly selected paragraphs (Rand\ $\oplus \ z_0$), and the concatenated same-intrinsic-task paragraphs gathered using the retrieval method in PICL \textit{before filtering} ($R(z_0)\oplus z_0$). We can see that the data constructed by retrieval has much lower perplexity and correspondingly yields higher accuracy scores, which verifies its usefulness. In Appendix \ref{sec:app_case}, we present several examples of the retrieved paragraphs and the corresponding intrinsic tasks.

\section{Related Work}

\paragraph{In-Context Learning} Recently, in-context learning (ICL), where models perform tasks simply conditioning on instructions or the concatenation of examples in the context~\cite{gpt3}, has been found promising for using PLMs in various application scenarios. To this end, there emerge many works to improve the ICL performance by calibrating the model predictions~\cite{calibrate,conca,surface-form-competition,noisy-channel-icl}, selecting or reordering demonstrations~\cite{retrieval-icl,what-make-good-example,prompt_order}, designing pre-training tasks~\cite{self-sup}, and breaking the context length limits~\cite{structured-prompting}. However, the underlying mechanism of ICL is poorly understood~\cite{rethinking-demons}. Therefore, some works propose mathematical frameworks to reveal how ICL works~\cite{bayesian-icl,induction-heads,transformer-circuits}, or investigate the pre-training data to explain ICL's good performance~\cite{data-distribution-icl,pre-training-corpora-icl}.

\paragraph{Multi-Task Fine-tuning for Cross-Task Generalization}
Fine-tuning PLMs on a large collection of downstream tasks enables generalization to unseen tasks under zero-shot~\cite{flan,t0,instruct-gpt,flan-t5} and few-shot ~\cite{metaicl,meta-in-context-tuning,natural-instructions,icl-function-class} scenarios. However, the performance of multi-task fine-tuning is largely restricted by the diversity of the annotated training tasks~\cite{udit}, which requires massive human efforts to scale up. In addition, direct training on downstream tasks easily brings undesired bias. In this work, we propose to meta-train the model with the intrinsic tasks automatically collected from the large-scale general corpus, which is easier to scale up and introduces little bias.

\paragraph{Pre-training Data Programming}
The conventional pre-training paradigm trains the model on plain-text corpora with the language modeling objective~\cite{gpt,gpt2,gpt3}. Recently works have found that carefully designed pre-training instances can further boost specific abilities like prompt adaption~\cite{ppt}, reasoning~\cite{term-freq-for-reason}, or sentence representation~\cite{icl-inductive-bias}. 
Our work studies constructing pre-training instances to improve the PLM's ICL ability while still maintaining its generalization on various NLP tasks.

\section{Conclusion}

This paper presents PICL, a framework that exploits the in-context learning ability of PLMs by pre-training models on concatenations of text paragraphs sharing the same ``intrinsic tasks'' gathered from the large-scale general corpus. In PICL, models learn to perform various intrinsic tasks conditioning on their context while preserving their generalization due to the little bias of the pre-training data. Extensive experiments show that PICL improves the ICL performance on various datasets against several baselines, enabling a 770 M model to outperform a larger model with about 4x parameters while maintaining good generalization across a wide range of tasks. For future work, we would like to consider adding human instructions to our pre-training framework to enhance more abilities of PLMs like zero-shot instruction following.

% 1. PICL怎么做的 2. PICL work 的机制 3. 实验的发现 4. 未来工作

%The pre-training data are constructed by gathering same-intrinsic-task paragraphs from a large-scale general corpus with a simple retrieval-based method and has little bias on input formats, label spaces, and data domains.
%to learn from its context to perform various ``intrinsic tasks'' naturally occurred in text paragraphs. We find that paragraphs with the same intrinsic tasks can be gathered from a large-scale general corpus with a simple retrieval-based method and the pre-training instances constructed by these paragraphs contain little bias on the input format, label spaces, and data domain. 

\section*{Limitations}

One limitation of our paper is that the exact distribution of the intrinsic tasks in the original corpus and the constructed data is still unknown. Knowing the distribution can offer a better interpretation of the effectiveness of PICL, even of the strong performance of large language models. Besides, although we can find many constructed instances that share obvious intrinsic tasks (see Appendix \ref{sec:app_case}), there still exist some instances where the intrinsic tasks are hard to identify. How to better evaluate the contribution of these instances to the ICL ability or designing better filtering approaches to select more informative data for ICL is worth studying.

Our task-semantics encoder inevitably contains some bias because it is trained on downstream datasets, although we have tried to ensure a large number and diversity of the dataset collection. However, the final language model is pre-trained on the general corpus, and we add the full document loss, which eliminates the bias to some extent.

Regarding computing power, we acknowledge that our framework takes relatively large training resources in the retrieval and pre-training process. Therefore, we did not conduct experiments based on extra-large language models.

\section*{Acknowledgements}
This work was supported by the NSFC projects (Key project with No. 61936010 ). This work was also supported by the Guoqiang Institute of Tsinghua University, with Grant No. 2020GQG0005.

\bibliography{custom}
\bibliographystyle{acl_natbib}

\clearpage

\appendix
% \section*{Appendices}
% \section*{Apendices}

\section{Details of the Pre-training Corpus}
\label{sec:app_pretrain_corpus}

This section presents details of the data processing of the pre-training corpus and its statistics.
\paragraph{Data Processing} Our pre-training corpus is a merge of \textsc{OpenWebText}~\cite{openwebtext}, \textsc{WikiCorpus}~\cite{wikidump}, and \textsc{BookCorpus}~\cite{bookcorpus}, downloaded from the HuggingFace datasets repository\textsuperscript{\ref{hf_data}}. We first split each document in the corpus into paragraphs with ``$\backslash$n''. To avoid training with too short paragraphs, we concatenate a paragraph with previous paragraphs if the token number after concatenation is lower than 128. We also exclude paragraphs longer than 500 tokens because they are not likely to fit into an instance with more than 1 paragraph. The filtering process in Section \ref{sec:pre-train} drops about 24\% instances. The licenses of all corpora allow for scientific research. 

\begin{figure}[t]
    \subfigbottomskip=0cm
    \subfigure[Paragraph Length] { \label{fig:len_mean} 
    \centering
    \includegraphics[width=0.475\linewidth]{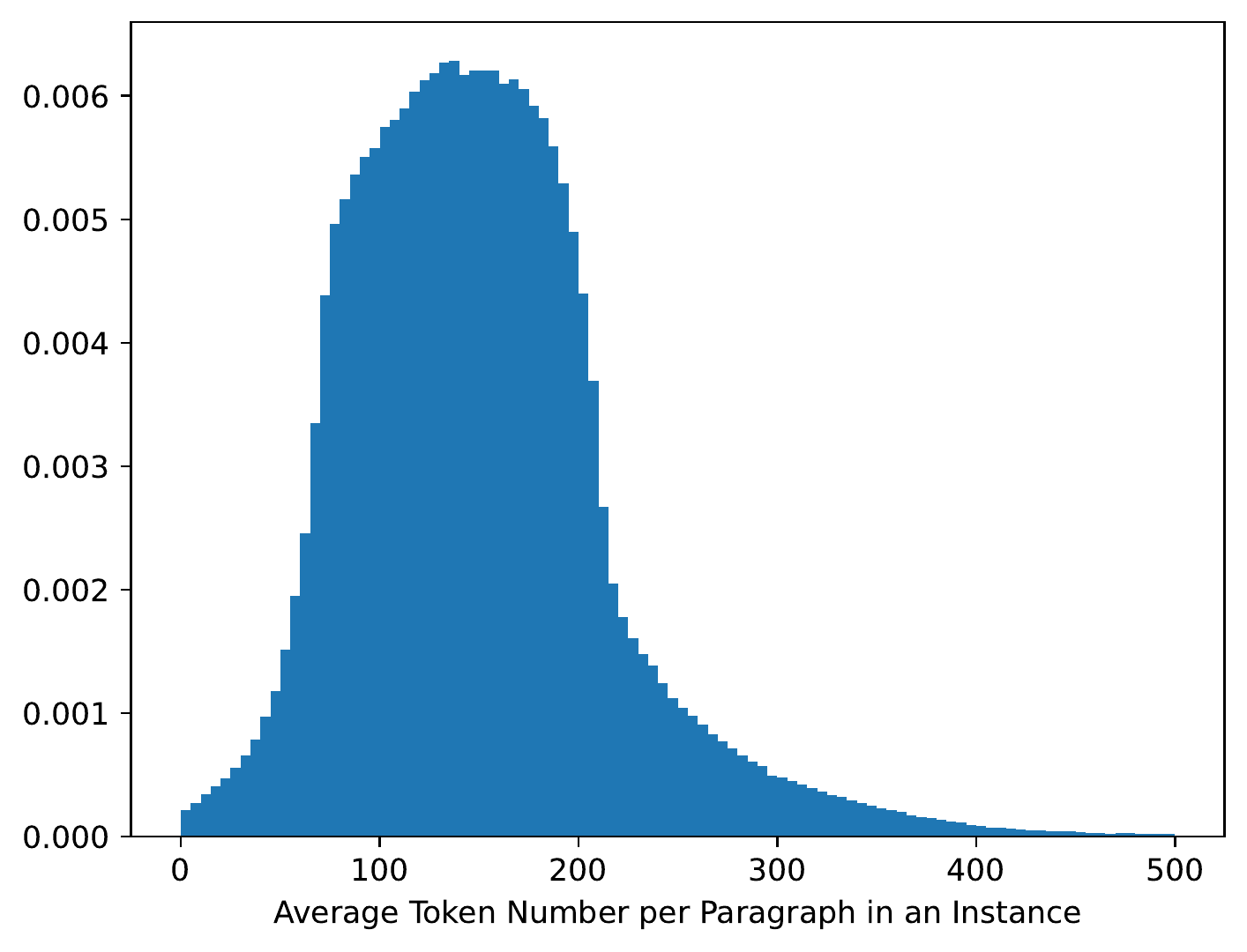} 
    } 
    \subfigure[Paragraph Number] { \label{fig:stat_shot} 
    \centering
    \includegraphics[width=0.465\linewidth]{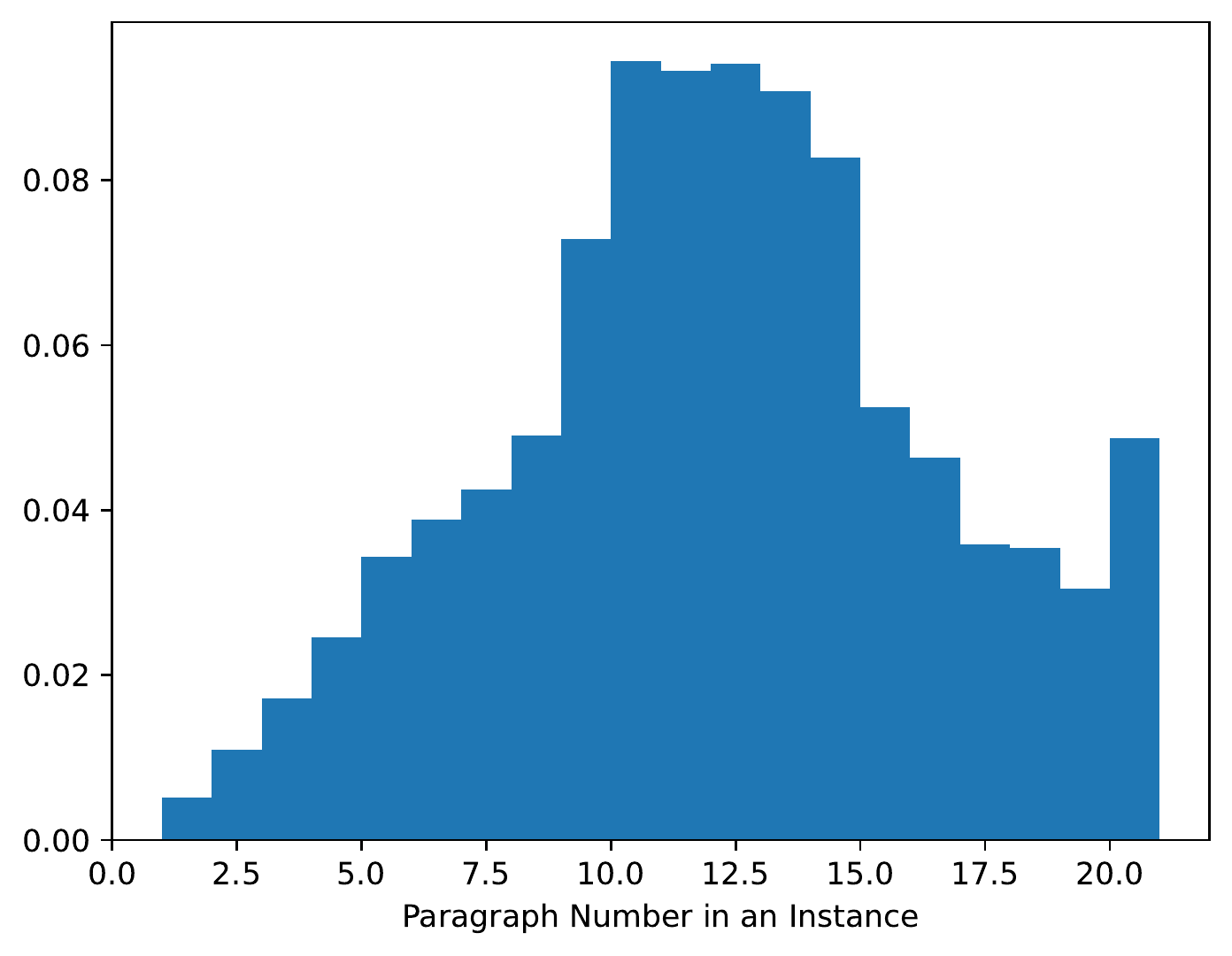} 
    }
    \caption{Pre-training data statistics. \textbf{(a)}: the distribution of the average paragraph length per instance. \textbf{(b)}: the distribution of the paragraph number per instance.} 
\end{figure}

\paragraph{Statistics} We plot the distribution of the mean paragraph length per instance in Figure \ref{fig:len_mean} and the distribution of the paragraph number per instance in Figure \ref{fig:stat_shot}. The average paragraph length is 150.0, and the average paragraph number in an instance is 11.7. We can see that the model sees various demonstration numbers in PICL pre-training.

\section{Details of the Evaluation Data}
\subsection{Few-shot Text Classification}
\label{sec:app_eval_data_detail}
The details of each text classification dataset and the corresponding prompt in evaluation are listed in Table \ref{tab:data_tc}. All datasets are downloaded from the HuggingFace datasets repository\textsuperscript{\ref{hf_data}}. We simplify the evaluation prompts as much as possible to reduce the effect of prompt engineering. Following previous works~\cite{gpt3,t0}, the model is evaluated by the \textit{ranking score} strategy, where we compare the perplexity of each classification label under the model and choose the label with the lowest perplexity. The licenses of all datasets allow for scientific research.

\subsection{Instruction Following}
\label{sec:app_sni}

The original test split of the benchmark \textsc{Super-NaturalInstructions}~\cite{super-natural-instructions} contains 119 tasks. We exclude tasks that appear in the training tasks of the task-semantics encoder or whose input length is too long to fit in the context of our model. Our final evaluation includes 105 tasks. A full list of the tasks is shown in Table \ref{tab:data_sni}. We use the same template to combine few-shot examples with task instructions as ~\citet{super-natural-instructions}. The license of this benchmark is Apache License 2.0.

\section{Details of the Downstream Training Data}
\label{sec:app_data_detail}
The downstream datasets we use to train the task-semantics encoder are a merge of the training data used in ~\cite{t0} and the HR$\rightarrow$LR setting in ~\cite{metaicl}. All datasets are downloaded from the HuggingFace datasets repository~\footnote{\url{https://huggingface.co/datasets/}\label{hf_data}} and all prompts come from the PromptSource library~\cite{p3}\footnote{\url{https://github.com/bigscience-workshop/promptsource}}. We exclude datasets from the sentiment classification task, the topic classification task, and the natural language inference task because they are included in our text classification evaluation. We finally get a collection of 37 datasets, as listed in Table \ref{tab:all_data}. The licenses of all datasets allow for scientific research.

\section{More Experimental Details}
All model checkpoints we used come from the HuggingFace models repository\footnote{\url{https://huggingface.co/models}}.  The searching interval of each hyper-parameter is listed in Table \ref{tab:hp}.

\begin{table}[t]
    \centering
    \small
    \begin{tabular}{llc}
    \toprule
    Stage & Name            & Values \\
    \midrule
    \multirow{3}{*}{Retrieve}
    &  learning rate   & 1e-4, \textbf{5e-5}, 1e-5 \\
    &  batch size   &   16, 32, \textbf{64} \\
    &  hard negatives & 0, 1, \textbf{4} \\
    \midrule
    \multirow{3}{*}{Pre-train}
    &  learning rate   & 1e-5, 5e-6, \textbf{1e-6}, 2e-7 \\
    &  batch size   &  64, \textbf{128}, 256, 512  \\
    &  warmup &  0, \textbf{1000}, 5000\\
    \bottomrule
    \end{tabular}
    \caption{Searching intervals of hyper-parameters.}
    \label{tab:hp}
\end{table}

\section{More Results}
\subsection{Results on Larger Base Model}
\label{app:large_model}
We test PICL based on the GPT2-xLarge~\cite{gpt2} with 1.5B parameters. From the results in Figure \ref{tab:icl_xl}, we can see that PICL is also applicable to larger models, outperforming the baselines based on the same-sized model on most datasets.

\subsection{Instruction Following}
\label{app:res_nat}
\begin{figure*}
    \centering
    \includegraphics[height=0.8\textheight]{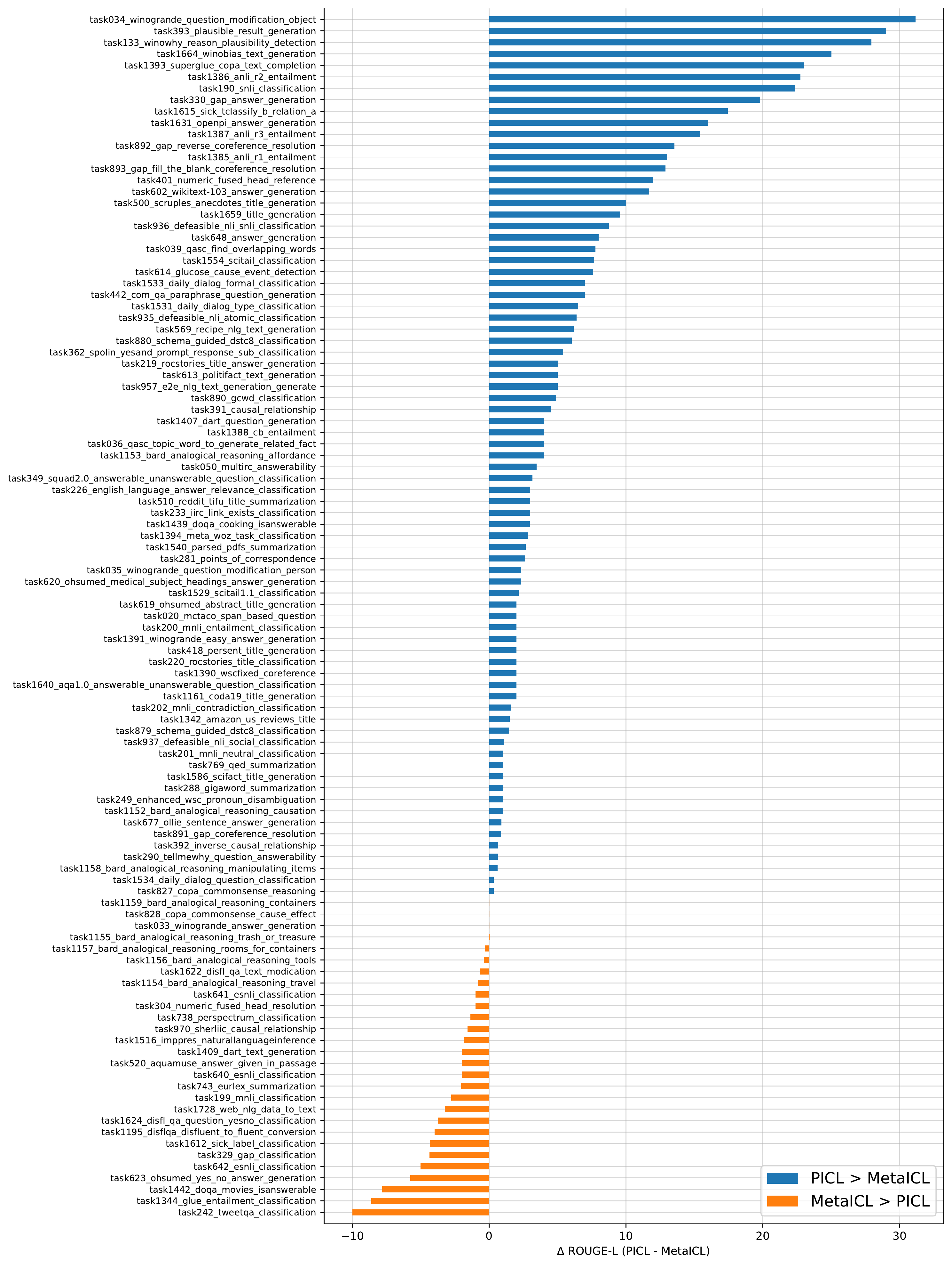}
    \caption{Per-task results of the comparison between PICL and MetaICL.}
    \label{fig:sni_full}
\end{figure*}

We present the performance comparison between PICL and MetaICL per evaluation task in Figure \ref{fig:sni_full}. PICL outperforms MetaICL on 77 / 105 tasks, indicating that PICL ensures the better generalization of the trained model. The name of each task is also listed in Figure \ref{fig:sni_full}. We can see that the top three tasks where MetaICL performs the best are:
\begin{itemize}[noitemsep,topsep=5pt]
    \small
    \item \texttt{doqa\_movies\_isanswerable},
    \item \texttt{glue\_entailment\_classification},
    \item \texttt{tweetqa\_classification},
\end{itemize}
which are all ``Yes/No'' classification tasks. The top three tasks where PICL performs the best are:
\begin{itemize}[noitemsep,topsep=5pt]
    \small
    \item \texttt{winogrande\_question\_modification\_object},
    \item \texttt{plausible\_result\_generation},
    \item \texttt{winowhy\_reason\_plausibility\_detection} ,
\end{itemize}
which are text generation, text generation, and ``Correct/Wrong'' classification tasks respectively. The four tasks where PICL and MetaICL have the same scores are:
\begin{itemize}[noitemsep,topsep=5pt]
    \small
    \item \texttt{bard\_analogical\_reasoning\_containers},
    \item \texttt{copa\_commonsense\_cause\_effect},
    \item \texttt{winogrande\_answer\_generation},
    \item \texttt{bard\_analogical\_reasoning\_trash\_or\_treasure},
\end{itemize}
which belong to commonsense reasoning and word analogy tasks.
\section{Case Studies}
\label{sec:app_case}
In Table \ref{tab:cases} and \ref{tab:cases2}, we present several cases of the retrieved paragraphs and the corresponding intrinsic tasks. We can see that there exists a large range of intrinsic tasks in the constructed data and many of them do not appear in the training data of the task-semantics encoder, which shows the generalization of the encoder.

\begin{table*}[t]
    \centering
    \small
    \begin{tabular}{@{}p{4.3cm}p{6.4cm}p{4.3cm}@{}}
    \toprule
    COS-E~\cite{cos-e} & DREAM~\cite{dream} & QuAIL~\cite{quail} \\
    QuaRTz~\cite{quartz} & Social-IQA~\cite{social_i_qa} & WiQA~\cite{wiqa} \\
    CosmosQA~\cite{cosmos_qa} & QASC~\cite{qasc} & QUAREL~\cite{quarel} \\
    SciQ~\cite{sciq} & Wiki-Hop~\cite{wikihop} & Adversarial-QA~\cite{adversarial_qa} \\
    Quoref~\cite{quoref} & ROPES~\cite{ropes} & DuoRC~\cite{duorc}\\
    Hotpot-QA~\cite{hotpotqa} & Wiki-QA~\cite{wikiqa} & Common-Gen~\cite{commongen}\\
    Wiki-Bio~\cite{wiki_bio} & SAMSum~\cite{samsum} & XSum~\cite{xsum} \\
    MRPC~\cite{mrpc} & PAWS~\cite{paws} & QQP~\cite{qqp} \\
    art~\cite{art} & circa~\cite{circa} & discovery~\cite{discovery} \\
    Freebase\_QA~\cite{freebaseqa} & google\_wellformed\_query~\cite{google_wellformed_query} & HellaSwag~\cite{hella_swag} \\
    liar~\cite{liar} & piqa~\cite{piqa} & scitail~\cite{scitail} \\
    swag~\cite{swag} & tab\_fact~\cite{tabfact} & yahoo\_answer\_topics\footnote{https://webscope.sandbox.yahoo.com/catalog.php?datatype=l} \\
    DBpedia~\cite{dbpedia} & & \\
    \bottomrule
    \end{tabular}
    \caption{The references of the 37 downstream datasets used to train the task-semantics encoder.}
    \label{tab:all_data}
\end{table*}

\begin{table*}[t]
    \centering
    \small
    \begin{tabular}{l|l|l|p{3.8cm}}
    \toprule
       Dataset  &  Task & Prompt & Label Space\\
    \midrule
       SST2     &  Sent. CLS    &  Sentence: \{sentence\} Label: \{label\} & Negative / Positive    \\ \midrule
       SST5     &  Sent. CLS    &  Sentence: \{sentence\} Label: \{label\} & Terrible / Bad / Neutral / Good / Great     \\ \midrule
       MR       &  Sent. CLS    &  Sentence: \{sentence\} Label: \{label\} & Negative / Positive     \\ \midrule
       RTE      &  NLI  &  Passage: \{premise\} Question: \{hypothesis\} Answer: \{label\}  &  Yes / No     \\ \midrule
       CB       &  NLI  &  Passage: \{premise\} Question: \{hypothesis\} Answer: \{label\} &  Yes / No / Maybe\\ \midrule
       SUBJ     &  Subj. CLS & Input: \{text\} Type: \{label\} & Objective / Subjective \\ \midrule
       AgNews   &  Topic CLS        & Sentence: \{text\} Label: \{label\}  &  World politics / Sports / Business / Science and technology\\
    \bottomrule
                
    \end{tabular}
    \caption{Details of the text classification datasets. ``Sent. CLS'' stands for ``Sentiment Classification''. ``NLI'' stands for ``Natural Language Inference''. ``Subj. CLS'' stands for ``Subjectivity Classification''. ``Topic CLS'' stands for ``Topic Classification''.}
    \label{tab:data_tc}
\end{table*}

\begin{table*}[t]
\small
\centering
\begin{tabular}{llccccccc|c}
\toprule
 % \multirow{2}{*}{Method}            & \multirow{2}{*}{Param.} & \multicolumn{8}{c}{\textit{Text Classification}} \\
% \cmidrule(lr){3-10}
Shot & Method                         & SST2                 & SUBJ                  & MR                    & RTE                   & AgNews                    & CB                    & SST5          & Average\\ \midrule
\multirow{3}{*}{GPT-xlarge}
& VanillaICL                          & 74.9$_{9.7}$         & 65.2$_{10.0}$         & 61.9$_{6.5}$          & 50.4$_{0.4}$          & 65.6$_{4.8}$              & 67.8$_{5.6}$          & 32.4$_{4.6}$ & 59.7$_{2.4}$\\
% & ExtraLM                             & 75.4$_{8.4}$         & 63.3$_{6.5}$          & 66.2$_{7.3}$          & 50.9$_{0.3}$          & 63.8$_{6.2}$              & 66.7$_{2.9}$          & 33.0$_{1.8}$ & 59.9$_{2.2}$\\
% % Random                              & 73.6$_{11.8}$        & 56.9$_{6.4}$          & \underline{73.0$_{9.0}$}      & 50.8$_{0.7}$      & 59.3$_{3.3}$      & 67.6$_{3.4}$      & 28.4$_{3.8}$ & 58.5$_{2.4}$\\
% & Self-Sup                            & 64.9$_{5.5}$        & 59.5$_{4.4}$          & 65.7$_{8.7}$         & 52.6$_{1.3}$          & 60.9$_{3.5}$             & 60.4$_{9.0}$         &  35.3$_{2.8}$ & 57.0$_{1.2}$ \\ 
& MetaICL                             & 71.1$_{2.0}$         & 64.9$_{7.6}$          & 66.8$_{6.3}$          & \textbf{60.0$_{2.8}$} & 66.2$_{5.4}$              & 64.4$_{1.6}$          &  34.6$_{3.7}$ & 61.2$_{1.3}$ \\ 
\cmidrule(l){2-10}
& PICL                                & \textbf{86.9$_{2.8}$} & \textbf{72.5$_{7.3}$} & \textbf{76.2$_{4.6}$} & 54.0$_{2.7}$         & \textbf{67.1$_{6.0}$}     & \textbf{70.0$_{4.6}$}          & \textbf{38.0$_{4.2}$} & \textbf{66.4$_{1.6}$}\\
\midrule
\multirow{3}{*}{GPT-\textit{Neo}}
& VanillaICL                          & 75.0$_{7.5}$        & 65.4$_{2.9}$          & 71.4$_{13.3}$          & 49.8$_{1.8}$          & 65.6$_{2.8}$              & 60.0$_{2.1}$          & 32.1$_{5.4}$ & 59.9$_{1.2}$\\
% & ExtraLM                             & 72.1$_{4.9}$         & 54.9$_{4.8}$          & 66.6$_{5.9}$          & 52.8$_{3.8}$          & 60.0$_{9.1}$              & 68.1$_{6.1}$          & 35.0$_{2.8}$ & 58.5$_{2.5}$\\
% % Random                              & 73.6$_{11.8}$        & 56.9$_{6.4}$          & \underline{73.0$_{9.0}$}      & 50.8$_{0.7}$      & 59.3$_{3.3}$      & 67.6$_{3.4}$      & 28.4$_{3.8}$ & 58.5$_{2.4}$\\
% & Self-Sup                            & 68.2$_{63.8}$       & 63.8$_{4.1}$          & 66.5$_{5.7}$         & 51.8$_{1.0}$         & 56.4$_{1.7}$              & 64.1$_{1.6}$          & 35.5$_{2.2}$ & 58.1$_{2.0}$ \\ 
& MetaICL                             & 80.1$_{5.8}$         & 55.6$_{9.5}$          & 73.1$_{9.0}$          & \textbf{57.5$_{3.9}$} & 64.2$_{3.4}$              & \textbf{65.5$_{6.4}$}          &  32.8$_{4.7}$ & 61.3$_{1.4}$ \\ 
\cmidrule(l){2-10}
& PICL                                & \textbf{86.4$_{1.0}$} & \textbf{68.6$_{5.7}$} & \textbf{83.6$_{2.4}$} & 50.2$_{0.7}$        & \textbf{67.5$_{1.2}$}     & 63.1$_{3.7}$ & \textbf{35.7$_{3.6}$} & \textbf{65.0$_{1.1}$}\\

\bottomrule
\end{tabular}
\caption{4-shot text classification results based on GPT2-XL (1.5B) and GPT-\textit{Neo} (2.7B). We report the average accuracy scores and the standard deviations across 5 random seeds for selecting demonstrations. The best scores on each dataset are in \textbf{boldface}.}
\label{tab:icl_xl}
\end{table*}

\begin{table*}[t]
    \centering
    \scriptsize
    \begin{tabular}{@{}p{2.8cm}|p{5.5cm}p{6.7cm}@{}}
    \toprule
    Task Category                   & \multicolumn{2}{c}{List of Tasks} \\ \midrule
\multicolumn{3}{c}{\textit{Evaluation Tasks} (105)} \\
    \midrule
    Coreference Resolution         
& task893\_gap\_fill\_the\_blank\_coreference\_resolution & task1664\_winobias\_text\_generation \\
& task648\_answer\_generation & task304\_numeric\_fused\_head\_resolution \\
& task891\_gap\_coreference\_resolution & task033\_winogrande\_answer\_generation \\
& task892\_gap\_reverse\_coreference\_resolution & task401\_numeric\_fused\_head\_reference \\
& task1390\_wscfixed\_coreference & task133\_winowhy\_reason\_plausibility\_detection \\
& task330\_gap\_answer\_generation & task329\_gap\_classification \\
& task249\_enhanced\_wsc\_pronoun\_disambiguation & task1391\_winogrande\_easy\_answer\_generation \\
\midrule
    Textual Entailment              
& task641\_esnli\_classification & task1529\_scitail1.1\_classification \\
& task202\_mnli\_contradiction\_classification & task1344\_glue\_entailment\_classification \\
& task1387\_anli\_r3\_entailment & task738\_perspectrum\_classification \\
& task890\_gcwd\_classification & task1612\_sick\_label\_classification \\
& task936\_defeasible\_nli\_snli\_classification & task1386\_anli\_r2\_entailment \\
& task201\_mnli\_neutral\_classification & task1385\_anli\_r1\_entailment \\
& task1516\_imppres\_naturallanguageinference & task1615\_sick\_tclassify\_b\_relation\_a \\
& task970\_sherliic\_causal\_relationship & task199\_mnli\_classification \\
& task935\_defeasible\_nli\_atomic\_classification & task937\_defeasible\_nli\_social\_classification \\
& task1388\_cb\_entailment & task1554\_scitail\_classification \\
& task190\_snli\_classification & task200\_mnli\_entailment\_classification \\
& task640\_esnli\_classification & task642\_esnli\_classification \\
\midrule
    Cause Effect Classification
& task1393\_superglue\_copa\_text\_completion & task391\_causal\_relationship \\
& task828\_copa\_commonsense\_cause\_effect & task614\_glucose\_cause\_event\_detection \\
& task827\_copa\_commonsense\_reasoning & task393\_plausible\_result\_generation \\
& task392\_inverse\_causal\_relationship &  \\
\midrule
    Title Generation
& task288\_gigaword\_summarization & task1161\_coda19\_title\_generation \\
& task619\_ohsumed\_abstract\_title\_generation & task500\_scruples\_anecdotes\_title\_generation \\
& task569\_recipe\_nlg\_text\_generation & task1586\_scifact\_title\_generation \\
& task602\_wikitext-103\_answer\_generation & task769\_qed\_summarization \\
& task510\_reddit\_tifu\_title\_summarization & task743\_eurlex\_summarization \\
& task1342\_amazon\_us\_reviews\_title & task418\_persent\_title\_generation \\
& task220\_rocstories\_title\_classification & task1659\_title\_generation \\
& task219\_rocstories\_title\_answer\_generation & task1540\_parsed\_pdfs\_summarization \\
\midrule
    Dialogue Act Recognition
& task880\_schema\_guided\_dstc8\_classification & task1531\_daily\_dialog\_type\_classification \\
& task1394\_meta\_woz\_task\_classification & task362\_spolin\_yesand\_prompt\_response\_sub\_classification \\
& task1533\_daily\_dialog\_formal\_classification & task879\_schema\_guided\_dstc8\_classification \\
& task1534\_daily\_dialog\_question\_classification &  \\
\midrule
    Answerability Classification
& task1439\_doqa\_cooking\_isanswerable & task1640\_aqa1.0\_answerable\_unanswerable\_question\_classification \\
& task242\_tweetqa\_classification & task1442\_doqa\_movies\_isanswerable \\
& task233\_iirc\_link\_exists\_classification & task290\_tellmewhy\_question\_answerability \\
& task520\_aquamuse\_answer\_given\_in\_passage & task226\_english\_language\_answer\_relevance\_classification \\
& task050\_multirc\_answerability & task349\_squad2.0\_answerable\_unanswerable\_question\_classification \\
& task1624\_disfl\_qa\_question\_yesno\_classification & task020\_mctaco\_span\_based\_question \\
\midrule
    Data to Text
& task1728\_web\_nlg\_data\_to\_text & task1409\_dart\_text\_generation \\
& task1407\_dart\_question\_generation & task957\_e2e\_nlg\_text\_generation\_generate \\
& task677\_ollie\_sentence\_answer\_generation & task1631\_openpi\_answer\_generation \\
\midrule
    Keyword Tagging
& task036\_qasc\_topic\_word\_to\_generate\_related\_fact & task620\_ohsumed\_medical\_subject\_headings\_answer\_generation \\
& task613\_politifact\_text\_generation & task623\_ohsumed\_yes\_no\_answer\_generation \\
\midrule
    Word Analogy
& task1159\_bard\_analogical\_reasoning\_containers & task1154\_bard\_analogical\_reasoning\_travel \\
& task1152\_bard\_analogical\_reasoning\_causation & task1155\_bard\_analogical\_reasoning\_trash\_or\_treasure \\
& task1156\_bard\_analogical\_reasoning\_tools & task1157\_bard\_analogical\_reasoning\_rooms\_for\_containers \\
& task1153\_bard\_analogical\_reasoning\_affordance & task1158\_bard\_analogical\_reasoning\_manipulating\_items \\
\midrule
    Overlap Extraction
& task039\_qasc\_find\_overlapping\_words & task281\_points\_of\_correspondence \\
\midrule
    Question Rewriting
& task035\_winogrande\_question\_modification\_person & task1195\_disflqa\_disfluent\_to\_fluent\_conversion \\
& task034\_winogrande\_question\_modification\_object & task442\_com\_qa\_paraphrase\_question\_generation \\
& task1622\_disfl\_qa\_text\_modication &  \\
\midrule
\multicolumn{3}{c}{\textit{Excluded Tasks} (14)} \\ \midrule
& task1356\_xlsum\_title\_generation & task670\_ambigqa\_question\_generation \\
& task645\_summarization & task760\_msr\_sqa\_long\_text\_generation \\
& task402\_grailqa\_paraphrase\_generation & task1598\_nyc\_long\_text\_generation \\
& task671\_ambigqa\_text\_generation & task121\_zest\_text\_modification \\
& task1345\_glue\_qqp\_question\_paraprashing & task1557\_jfleg\_answer\_generation \\
& task232\_iirc\_link\_number\_classification & task1358\_xlsum\_title\_generation \\
& task1562\_zest\_text\_modification & task102\_commongen\_sentence\_generation \\
     \bottomrule
    \end{tabular}
    \caption{A full list of the evaluation tasks we use and the tasks we exclude from the original test split of \textsc{Super-NaturalInstructions}~\cite{super-natural-instructions}.}
    \label{tab:data_sni}
\end{table*}

\begin{table*}[t]
    \centering
    \small
    \begin{tabular}{@{}cp{11.5cm}p{3cm}@{}}
    \toprule
    ID & Paragraphs                                    & Intrinsic Task \\ \midrule
    1 & \begin{itemize}[leftmargin=*,nolistsep,noitemsep,topsep=0pt]
        \vspace{-5pt}
        \item Marko Jovanovski Marko Jovanovski (born 24 July 1988) is a Macedonian professional footballer who plays as a goalkeeper for Akademija Pandev.
        \item Andreas Paraskevas Andreas Paraskevas (; born 15 September 1998) is a Cypriot footballer who plays as a goalkeeper for Doxa Katokopias.
        \item Evripidis Giakos Evripidis Giakos (; born 9 April 1991) is a Greek professional footballer who plays as an attacking midfielder for Super League 2 club AEL.
        \vspace{-8pt}
        \end{itemize} &  World Knowledge Completion\\ \midrule
    2 & \begin{itemize}[leftmargin=*,nolistsep,noitemsep,topsep=0pt]
        \vspace{-5pt}
        \item The Hive scouting teams had been infiltrating our space for several weeks, sending three- and six-man teams in. In short, they were making me look bad on the home world.
        \item And for good measure, Walker ordered the Wisconsin National Guard to prepare to intervene in case of any strike action by unions. In a word, Walker wants the destruction of organized labor in Wisconsin.
        \item "Scientists began examining him... he was covered in tattoos consisting of lines and dots,... 80 percent of the points correspond to those used in acupuncture today." This means the Prince of Wales ought to start listening to scientists.
        \vspace{-8pt}
        \end{itemize} &  Intent Identification \\ \midrule
    3 & \begin{itemize}[leftmargin=*,nolistsep,noitemsep,topsep=0pt]
        \vspace{-5pt}
        \item ln(x) $\approx$ $\pi$ 2 M (1,2\^{}2 - m / x ) - m ln(2). \{$\backslash$displaystyle $\backslash$ln(x)$\backslash$approx \{$\backslash$frac \{$\backslash$pi \}\{2M(1,2\^{}\{2-m\}/x)\}\}-m$\backslash$ln(2).\}
        \item sin( x ) + 1 3 sin ( 3 x ) + 1 5 sin ( 5 x ) + $\cdots$. \{$\backslash$displaystyle $\backslash$sin(x)+\{$\backslash$frac \{1\}\{3\}\}$\backslash$sin(3x)+\{$\backslash$frac \{1\}\{5\}\}$\backslash$sin(5x)+$\backslash$dotsb.\}
        \item c q ( n ) = $\Sigma$ d | q $\mu$ ( q d ) $\eta$ d ( n ). \{$\backslash$displaystyle c\_\{q\}(n)=$\backslash$sum\_\{d$\backslash$mid q\}$\backslash$mu $\backslash$left(\{$\backslash$frac\{q\}\{d\}\}$\backslash$right)$\backslash$eta\_\{d\}(n).\}
        \vspace{-8pt}
        \end{itemize} &  Latex Equation Translation \\ \midrule
    4 & \begin{itemize}[leftmargin=*,nolistsep,noitemsep,topsep=0pt]
        \vspace{-5pt}
        \item How did Japan stumble on for another nine years, borrowing trillions of yen and squandering those trillions on make-work bridges to nowhere and lavish social spending? Answer: its citizens self-funded its deficits by saving trillions and investing those trillions in government debt.
        \item How did our country thrive without income taxes for 126 years? Answer: federal spending was significantly lower than it is today. In the early 1900s, government spending accounted for roughly 7\% of our GDP; today, federal spending accounts for around 35\% of our GDP.
        \item What was Trump’s biggest persuasion problem in the election? Answer: His opponents did a great job of framing him as some kind of Hitler.
        \vspace{-8pt}
        \end{itemize} &  Question Answering \\ \midrule
    5 & \begin{itemize}[leftmargin=*,nolistsep,noitemsep,topsep=0pt]
        \vspace{-5pt}
        \item An isopycnal is a line of constant density. An isoheight or isohypse is a line of constant geopotential height on a constant pressure surface chart. Isohypse and isoheight are simply known as lines showing equal pressure on a map. Temperature and related subjects
        \item Once theory is applied to a mechanical design, physical testing is often performed to verify calculated results. Structural analysis may be used in an office when designing parts, in the field to analyze failed parts, or in laboratories where parts might undergo controlled failure tests. Thermodynamics and thermo-science
        \item Complex numbers often generalize concepts originally conceived in the real numbers. For example, the conjugate transpose generalizes the transpose, hermitian matrices generalize symmetric matrices, and unitary matrices generalize orthogonal matrices. In applied mathematics Control theory
        \vspace{-8pt}
        \end{itemize} &  Topic Classification \\ \midrule
    6 & \begin{itemize}[leftmargin=*,nolistsep,noitemsep,topsep=0pt]
        \vspace{-5pt}
        \item (speaking to Elder Fortie): Is this something you always wanted to do? ELDER FORTIE: Nope. It’s not. SEVERSON: So why are you here? ELDER FORTIE: Because the idea of having an empty seat in heaven troubles me. SEVERSON: Sister Waymith is from Sweet, Idaho.
        \item (speaking to Steve Allen): Are there any countries in particular that you’re really zeroing in on, you’d really like to make some inroads? ALLEN: Yeah, the United States of America, North America. We’d like to make more inroads here. SEVERSON: Inroads like the church has made south of the border. Mexico, in particular, has been fertile ground for Mormon missionaries.
        \item (to Elder Russell): Why are you learning Mandarin if you’re going to Canada? ELDER RUSSELL: I guess there’s a sizable population up there. I mean, everyone deserves to hear our message, so we’ll go worldwide wherever they are. SEVERSON: This group is leaving soon for Ukraine. First, they had to be considered worthy of serving a mission.
        \vspace{-8pt}
        \end{itemize} &  Dialogue in a Script \\ \bottomrule
    \end{tabular}
    \caption{Cases of the retrieved paragraphs and the corresponding intrinsic tasks.}
    \label{tab:cases}
\end{table*}

\begin{table*}[t]
    \centering
    \small
    \begin{tabular}{@{}cp{11.5cm}p{3cm}@{}}
    \toprule
    ID & Paragraphs                                    & Intrinsic Task \\ \midrule
    7 & \begin{itemize}[leftmargin=*,nolistsep,noitemsep,topsep=0pt]
        \vspace{-5pt}
        \item Now we can log into our Twilio account and set the Message Request URL to our sms route via ngrok: Try the app out by texting into your new Twilio number and you’ll get the response back. Displaying Our Messages We’re now passing our message to the arduino. The next step is to write the code that examines that message and displays it on our LCD. Let’s lay the foundation for our app: \# include < Wire. h > \# include "rgb\_lcd.h" rgb\_lcd lcd ; void setup () \{ Serial. begin ( 9600 ); // set up the LCD's number of columns and rows: lcd. begin ( 16, 2 ); lcd. setCursor ( 0, 1 ); // Print a message to the LCD. lcd. print ( "Ricky's Pager" ); delay ( 1000 ); \} void loop () \{ \} 
        \item Now we’re ready to track allocations. The first step is to “hijack” our 3 memory functions we defined in the first part (lines 4, 11 and 17): void* \_Malloc(tU32 Size, tU32 AllocType, const tChar* Desc, const tChar* File, tU32 Line) \{ void* Result = malloc(Size); RegisterAlloc(Result, Size, AllocType, Desc, File, Line); return Result; \} void* \_Realloc(void* Ptr, tU32 Size, const tChar* File, tU32 Line) \{ void* Result = realloc(Ptr, Size); UpdateAlloc(Ptr, Result, Size, File, Line); return Result; \} void \_Free(void* Ptr) \{ UnregisterAlloc(Ptr); return free(Ptr); \}
        \item Here we use the gulp.src API to specify our input files. One thing to note is that we need to specify a reporter for JSHint. I’m using the default reporter, which should be fine for most people. More on this can be found on the JSHint website. Compress Images Next, we’ll set up image compression: gulp. task ( 'images', function () \{ return gulp. src ('src/images/**/*' ). pipe ( imagemin (\{ optimizationLevel : 3, progressive : true, interlaced : true \})). pipe ( gulp. dest ( 'dist/assets/img' )). pipe ( notify (\{ message : 'Images task complete' \})); \});
        \vspace{-8pt}
        \end{itemize} &  Code Generation\\ \midrule
    8 & \begin{itemize}[leftmargin=*,nolistsep,noitemsep,topsep=0pt]
        \vspace{-5pt}
        \item Note: GP = Games played; W = Wins; L = Losses; T = Ties; OTL = Overtime loss;  
        \item SOL = Shootout loss; GF = Goals for; GA = Goals against; Pts = Points National Conference
        \item ERA = Earned run average; SO = Strikeouts; +/- = Plus/Minus; PIM = Penalty minutes; GS = Games Started;
        \vspace{-8pt}
        \end{itemize} &  Word Abbriviation \\ \midrule
    9 & \begin{itemize}[leftmargin=*,nolistsep,noitemsep,topsep=0pt]
        \vspace{-5pt}
        \item He strode toward her, barely slowly when he reached her. One arm slid around her waist and the other along her shoulders. She’d barely registered his touch before his mouth descended upon hers.
        \item He lifted his head and stared at her. His face paled, and for the first time she noticed a spattering of orange freckles on his nose and across his cheekbones. He didn’t speak. Just stared.
        \item He continued rocking her gently, steadily. Her body’s tremors calmed and her sobs quietened. He removed his white handkerchief from his trouser pocket, wiping her face. She didn’t look at him and kept her eyes lowered. 
        \vspace{-8pt}
        \end{itemize} &  Vivid Description \\ \midrule
    10 & \begin{itemize}[leftmargin=*,nolistsep,noitemsep,topsep=0pt]
        \vspace{-5pt}
        \item Even the sugary cereals, they said, are of nutritional value because they contain vitamins and minerals. Research shows that 40 percent of U.S. children consume their milk via cereal, said Sutherland of Kellogg. General Mills cites data from the Journal of the American Dietetic Association that says people who frequently eat cereal, including kids who eat sweetened ones, tend to have healthier body weights than those that don’t.
        \item Lead’s toxicity has long been known, and most of the uses that led to human exposure, like the manufacture of lead paint, have been banned for decades. Lead ammunition consumed only about 3 percent of the 6.4 million tons of lead used worldwide in 2000, according to a 2003 report by the Nordic Council of Ministers.
        \item One reason Tesla has pushed the technology so aggressively is that its battery packs store more than three times the energy of its competitors’ electric-car batteries. As a result, they require more power to charge quickly, says Arindam Maitra, a senior project manager at the Electric Power Research Institute.
        \vspace{-8pt}
        \end{itemize} &  Scientific Evidence Generation \\
    % 5 & \begin{itemize}[leftmargin=*,nolistsep,noitemsep,topsep=0pt]
    %     \vspace{-5pt}
    %     \item An isopycnal is a line of constant density. An isoheight or isohypse is a line of constant geopotential height on a constant pressure surface chart. Isohypse and isoheight are simply known as lines showing equal pressure on a map. Temperature and related subjects
    %     \item Once theory is applied to a mechanical design, physical testing is often performed to verify calculated results. Structural analysis may be used in an office when designing parts, in the field to analyze failed parts, or in laboratories where parts might undergo controlled failure tests. Thermodynamics and thermo-science
    %     \item Complex numbers often generalize concepts originally conceived in the real numbers. For example, the conjugate transpose generalizes the transpose, hermitian matrices generalize symmetric matrices, and unitary matrices generalize orthogonal matrices. In applied mathematics Control theory
    %     \vspace{-8pt}
    %     \end{itemize} &  Topic Classification \\ \midrule
        \bottomrule
    \end{tabular}
    \caption{Cases of the retrieved paragraphs and the corresponding intrinsic tasks.}
    \label{tab:cases2}
\end{table*}

\end{document}